\documentclass[10pt,twocolumn,letterpaper]{article}
\usepackage[table]{xcolor}
\usepackage{booktabs}

\usepackage{iccv}
\usepackage{times}
\usepackage{epsfig}
\usepackage{graphicx}
\usepackage{amsmath}
\usepackage{amssymb}
\usepackage[breaklinks=true,bookmarks=false]{hyperref}

\usepackage{multirow}
\usepackage{float}
\usepackage{pdfrender}
\usepackage{comment}
\usepackage{lipsum}  
\usepackage{subcaption} % Ensure this package is included
\usepackage{siunitx}
\definecolor{lightgray}{gray}{0.9}

\iccvfinalcopy % *** Uncomment this line for the final submission

\def\iccvPaperID{1382} % *** Enter the ICCV Paper ID here

\ificcvfinal\fi

\begin{document}

\title{Sparse-Dense Side-Tuner for efficient Video Temporal Grounding}

%%%%%%%%% AUTHORS - PLEASE UPDATE
\author{David Pujol-Perich, Sergio Escalera, Albert Clapés\\
Universitat de Barcelona and Computer Vision Center, Barcelona, Spain\\
{\tt\small \{david.pujolperich, sescalera, aclapes\}@ub.edu}}

\maketitle

\ificcvfinal\thispagestyle{empty}\fi

\setlength\abovedisplayskip{4pt}
\setlength\belowdisplayskip{4pt}

\begin{abstract}
Video Temporal Grounding (VTG) involves Moment Retrieval (MR) and Highlight Detection (HD) based on textual queries. For this, most methods rely solely on final-layer features of frozen large pre-trained backbones, limiting their adaptability to new domains. While full fine-tuning is often impractical, parameter-efficient fine-tuning --and particularly side-tuning (ST)-- has emerged as an effective alternative. However, prior ST approaches this problem from a frame-level refinement perspective, overlooking the inherent sparse nature of MR. To address this, we propose the Sparse-Dense Side-Tuner (SDST), the first anchor-free ST architecture for VTG. We also introduce the \textit{Reference-based Deformable Self-Attention}, a novel mechanism that enhances the context modeling of the deformable attention --a key limitation of existing anchor-free methods. Additionally, we present the first effective integration of InternVideo2 backbone into an ST framework, showing its profound implications in performance. Overall, our method significantly improves existing ST methods, achieving highly competitive or SOTA results on QVHighlights, TACoS, and Charades-STA, while reducing up to a $73\%$ the parameter count w.r.t. the existing SOTA methods. The code is publicly accessible at \url{https://github.com/davidpujol/SDST}.
\vspace{-0.5cm}
\end{abstract}

\section{Introduction}

% Intro of the field
Recently the field of Video Understanding has gained attention due to its potential in applications like search engines or recommendation systems. A key task in this field is Video Temporal Grounding (VTG), which localizes moments within videos based on textual descriptions. Typically, this involves doing both Moment Retrieval (MR)~\cite{gao2017tall,jang2023knowing, lei2021detecting,gordeev2024saliency,cao2024flashvtg} and Highlight Detection (HD)~\cite{badamdorj2021joint, hong2020mini, sun2014ranking,xu2021cross, han2024unleash}. More specifically, MR predicts moment boundaries w.r.t. textual queries, while HD offers a more interpretable perspective, predicting frame-level saliency scores.

\begin{figure}[t]
\centering
\includegraphics[width=0.48\textwidth]{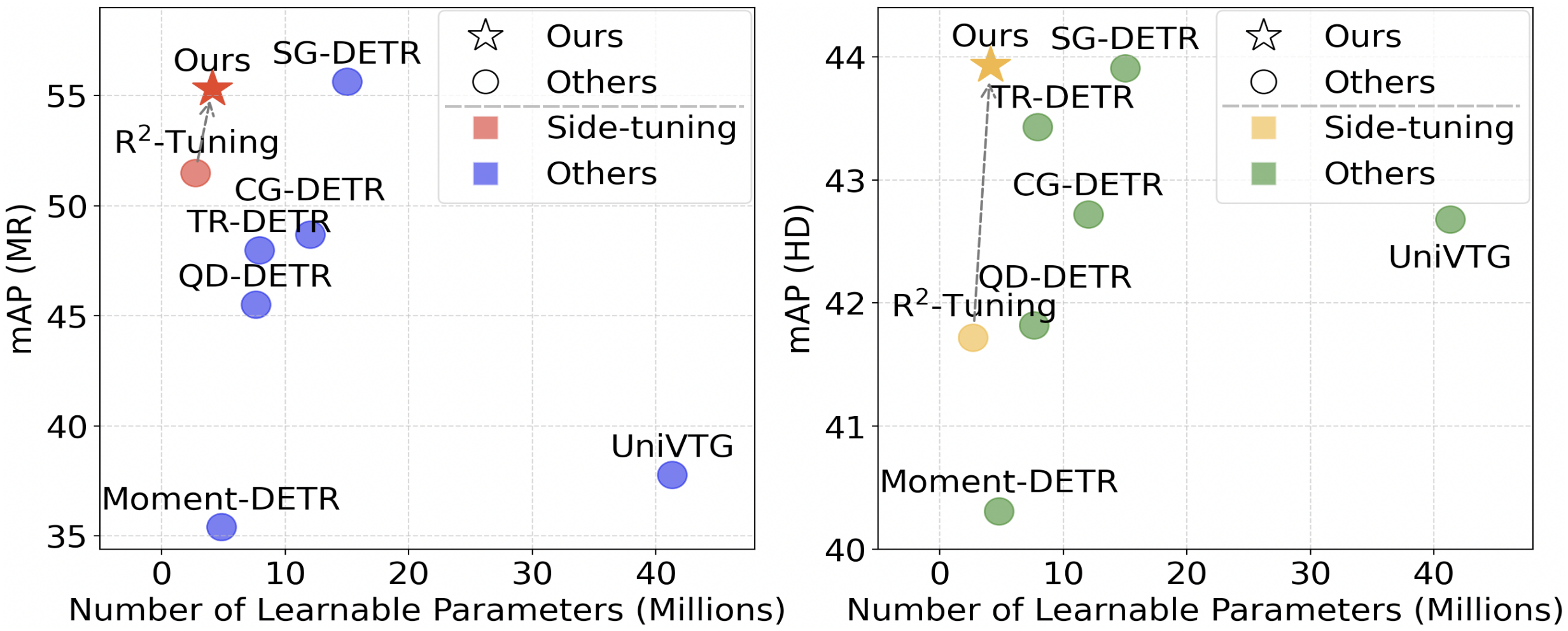}
\vspace{-0.7cm}
\caption{\label{fig:bubble_plot} Comparison of our proposed method and the main MR (left) and HD (right) baselines. These are all evaluated on QVHighlights \textit{val} split and use InternVideo2-1B features. These results show that our method improves existing side-tuning works and attains SOTA results while significantly reducing the number of trainable parameters.}
\vspace{-0.5cm}
\end{figure}

In the past, existing works focused on either MR or HD, given the lack of datasets that combined both tasks. QVHighlights~\cite{lei2021detecting} shifted this paradigm, proposing the first dataset suitable for this multi-task setup. Since then, various works have been proposed following \textit{anchor-based}~\cite{liu2022umt, lin2023univtg,cao2024flashvtg, liu2024r}, \textit{anchor-free}~\cite{lei2021detecting,moon2023correlation, gordeev2024saliency} and even LLM-based~\cite{lu2024llava, meinardus2024surprising} approaches. All of these, however, incur in a common critical limitation, which is relying on the final-layer features of frozen backbones. This is especially limiting when facing large distribution shifts between the pre-training distribution and that of the downstream task~\cite{pujol2025sada}. This issue is particularly pronounced in VTG, where an image-domain backbone~\cite{radford2021learning} is transferred to the video domain.

A typical solution is fine-tuning; however, full fine-tuning is impractical due to its high computational cost. Doing this more efficiently has thus become critical, motivating the interest in \textit{parameter-efficient fine-tuning (PEFT)}~\cite{han2024parameter} methods --e.g., Prompting or Adapters-- which optimize only a small subset of parameters. Unfortunately, these still require full back-propagation through the backbone, making it still memory-intensive. To address this, \cite{sung2022lst} introduced \textit{side-tuning} (ST), an effective PEFT, and \textit{memory-efficient fine-tuning} (MEFT) method. ST creates a parallel pathway to refine intermediate features, thus back-propagating over a minimal set of parameters only. In this regard, we highlight the work of R$^2$-Tuning~\cite{liu2024r}, the first ST approach for VTG, which recursively fuses multi-modal intermediate CLIP~\cite{radford2021learning} embeddings. \cite{liu2024r} also proposes adapting these frame-level embeddings for MR by generating a dense set of anchors. As shown in Fig.~\ref{fig:bubble_plot}, this method proves ineffective for MR -- a highly sparse task where videos may contain very few ground-truth action.

This motivates our dual-stream \textit{Sparse-Dense Side-Tuner (SDST)}, the first proposal-free ST method, carefully designed for multi-task learning across sparse (MR) and dense (HD) tasks. For this, SDST jointly refines frame-level embeddings --suitable for HD-- while learning what we call the \textit{recurrent decoder queries} for MR. We also identify an implicit contextual limitation of the deformable attention~\cite{zhu2020deformable} module of anchor-free architectures like ours, which in our task results in collapsing offsets around their initialization values. This also impedes the selection of keys outside of the current estimated moment boundaries, critical to potentially refine the boundaries of longer moments. Consequently, we propose the alternative \textit{Reference-based Deformable Self-Attention (RDSA)}, which naturally addresses this issue by reformulating the CA into a SA-based mechanism. Finally, we tackle the performance degradation of ST methods derived from the use of image-based CLIP, over more advanced spatio-temporal VLMs~\cite{wang2025internvideo2}, as noted in \cite{gordeev2024saliency}. A key challenge of exploiting \cite{wang2025internvideo2} is, however, defining an effective token pooling strategy, as naive strategies like CLS pooling yield significant performance drops. This motivates our proposed module re-utilization scheme, which results in the first successful integration of InternVideo2~\cite{wang2025internvideo2} into an ST framework.

% Summary of the contributions
In summary, our main contributions are threefold: 1) We propose SDST, the first anchor-free ST architecture, specifically tailored for complex sparse-dense multi-task setups like VTG. Our method significantly outperforms existing ST architectures on QVHighlights~\cite{lei2021detecting}, TACoS~\cite{rohrbach2014coherent} and Charades-STA~\cite{gao2017tall} (see Fig.~\ref{fig:bubble_plot}). SDST also performs competitively and even surpasses existing SOTA while reducing its learnable parameter count by up to $73\%$, and incurring in a minimal memory overhead. 2) We identify a key contextual limitation of the deformable attention mechanism that, in our task, results in collapsing offsets. Consequently, we propose RDSA, which tackles this issue by allowing more complex key-selection strategies. 3) We address the limitations of transfer learning from an image-domain backbone to a video domain by integrating, for the first time, the more advanced spatio-temporal backbone \cite{wang2025internvideo2} into an ST framework. This proves to be a non-trivial challenge with massive performance implications.

\vspace{-0.1cm}

\section{Related work}
\textbf{Video temporal grounding (VTG):} One relevant task in Video Understanding is VTG~\cite{lin2023univtg} which aims to identify actions from text queries. Traditionally, this task was approached either from a MR~\cite{gao2017tall, jang2023knowing, lei2021detecting, moon2023query, moon2023correlation, gordeev2024saliency, cao2024flashvtg} or HD~\cite{badamdorj2021joint, hong2020mini, sun2014ranking,xu2021cross, han2024unleash} perspective. MR focuses on predicting specific action proposals while HD aims to compute the saliency scores for each of the frames w.r.t. the query. Nevertheless, after the proposal of QVHighlights~\cite{lei2021detecting}, the literature started approaching this from a multi-task learning perspective. In this regard, arguably the most notable family of methods is that based on DETR~\cite{lei2021detecting, moon2023query, moon2023correlation, lee2025bam, gordeev2024saliency, xiao2024bridging} given the sparse nature of MR. These methods normally fuse both video and textual modality into a final embedding which is used for HD and as a memory of a Transformer decoder that refines a set of learnable queries. More in detail, Moment-DETR~\cite{lei2021detecting} constitutes the first baseline, which uses a standard Transformer encoder to process both modalities simultaneously. QD-DETR~\cite{moon2023query} proposes instead to inject the textual information to the video modality with cross attention (CA), followed by a temporal modeling module. CG-DETR~\cite{moon2023correlation} reduces the negative impact of irrelevant textual tokens by adding an Adaptive CA module that leverages dummy tokens to redirect attention weight. In this same line, SG-DETR~\cite{gordeev2024saliency} introduces a saliency-guided CA, weighting the contribution of irrelevant textual tokens based on the computed saliency scores. Despite the notable success of these proposal-free methods, one core limitation that they all share is that they rely solely on the last-layer features of a frozen CLIP~\cite{radford2021learning} and/or Slowfast~\cite{feichtenhofer2019slowfast} backbones. \cite{yan2023unloc} overcomes this limitation via the full fine-tuning of CLIP, even though this becomes nearly intractable given the resources that this requires. R$^2$-Tuning~\cite{liu2024r}, the most similar method to our work, overcomes this limitation by proposing the first PEFT and MEFT method for VTG, which recursively applies a multi-modal fusion ST. In this work, we argue that one of its core drawbacks, however, is its anchor-based nature, remaining oblivious to the shown benefits of DETR-based architectures in very sparse tasks like MR~\cite{liu2022dab, li2022dn, gordeev2024saliency}. This motivates our proposed SDST, which incorporates the best of both PEFT and MEFT.

\noindent \textbf{Parameter-and-memory-efficient fine-tuning:} Foundation models and VLMs~\cite{radford2021learning, wang2022internvideo} have become the corner-stone of recent advances in Video Understanding applications. The use of pre-extracted features remains, however, an important limitation to its application on downstream tasks like VTG~\cite{lin2023univtg}. Fine-tuning these huge models is often simply infeasible given the large resources that these require. This has motivated the rise of PEFT~\cite{han2024parameter}, which aims to democratize the fine-tuning process by restricting the tunable parameters to the bare minimum. In this regard, some of the most prominent approaches are Prompting~\cite{jia2022visual, zhou2022conditional} and the use of Adapters~\cite{pan2022st, gao2024clip}. The first keeps the backbone frozen while learning prompts that are appended to the input, bridging the gap between the backbone's expected distribution and that of the downstream task. Adapters, in contrast, incorporate small trainable modules in the backbone while keeping the rest unchanged. These methods, however, still require full backpropagation through the frozen model, making them memory-inefficient. Recently, the new paradigm of \textit{ST}~\cite{lin2022frozen, sung2022lst, qing2023disentangling} --and particularly \cite{liu2024r} for VTG-- has gained attraction creating a parallel pathway to exploit intermediate backbone representations while guaranteeing that backpropagation is only applied to this small parallel module. Interestingly, these methods normally rely on pre-extracted CLIP representations for tasks like HD/MR. As shown by \cite{gordeev2024saliency}, this is a very limiting aspect given the lack of temporal reasoning of CLIP or its difficulties in understanding textual queries beyond simple spatial descriptions. For this reason, in this work, we propose a novel ST architecture that, for the first time, relies on the more advanced InternVideo2~\cite{wang2025internvideo2} backbone. Unlike CLIP, this backbone is trained on video-domain inputs, resulting in enhanced spatial and temporal modeling capabilities.

\noindent \textbf{Deformable attention:} A key component of most SOTA methods for VTG like SG-DETR~\cite{gordeev2024saliency} is their deformable attention module~\cite{zhu2020deformable, xia2022vision}. This module addresses the well-studied slow convergence of DETR~\cite{carion2020end}. It does so by limiting the selectable keys based on the predictions of the learnable queries, completely decoupling both query and key spaces. Despite desirable in terms of efficiency, it has been previously noted in the literature~\cite{zhang2022dino} that the lack of context of the queries w.r.t. the key/value space can lead to suboptimal attendable key selection. VTG methods often mitigate this effect by partially or completely initializing the learnable queries based on the DETR-memory~\cite{zhang2022dino} --i.e., frame-level representations in our case. In this work, we empirically demonstrate the limitations of these initialization-based methods for VTG tasks, which motivates our proposed simple yet effective alternative, the \textit{Reference-based Deformable Self-Attention (RDSA)}. This reformulates the deformable CA into a deformable SA, naturally solving the aforementioned issue.

\vspace{-0.1cm}
\section{Method}

\begin{figure*}
    \centering
    \begin{subfigure}{0.63\textwidth} % 2/3 of the width
        \centering
        \includegraphics[width=\textwidth]{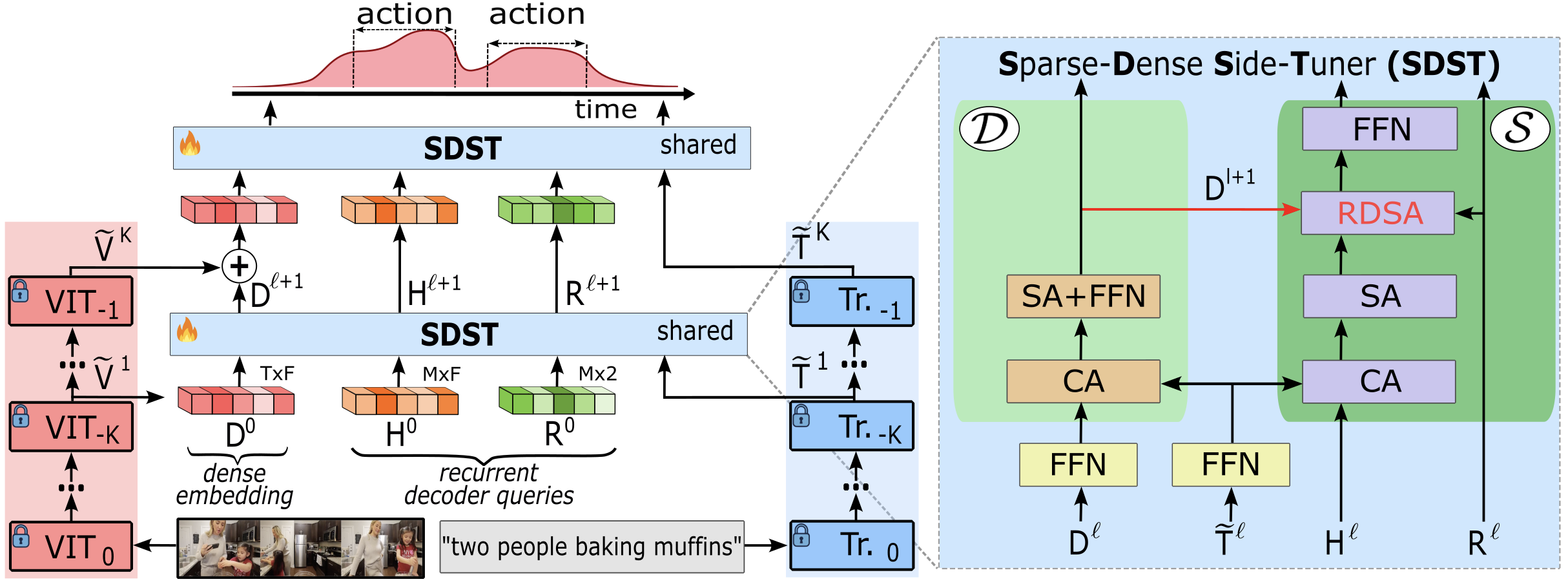}
        \vspace{-0.5cm}
        \caption{Overview of our proposed architecture.}
        \label{fig:main_architecture}
    \end{subfigure}
    \hfill
    \begin{subfigure}{0.33\textwidth} % 1/3 of the width
        \centering
        \includegraphics[width=\textwidth]{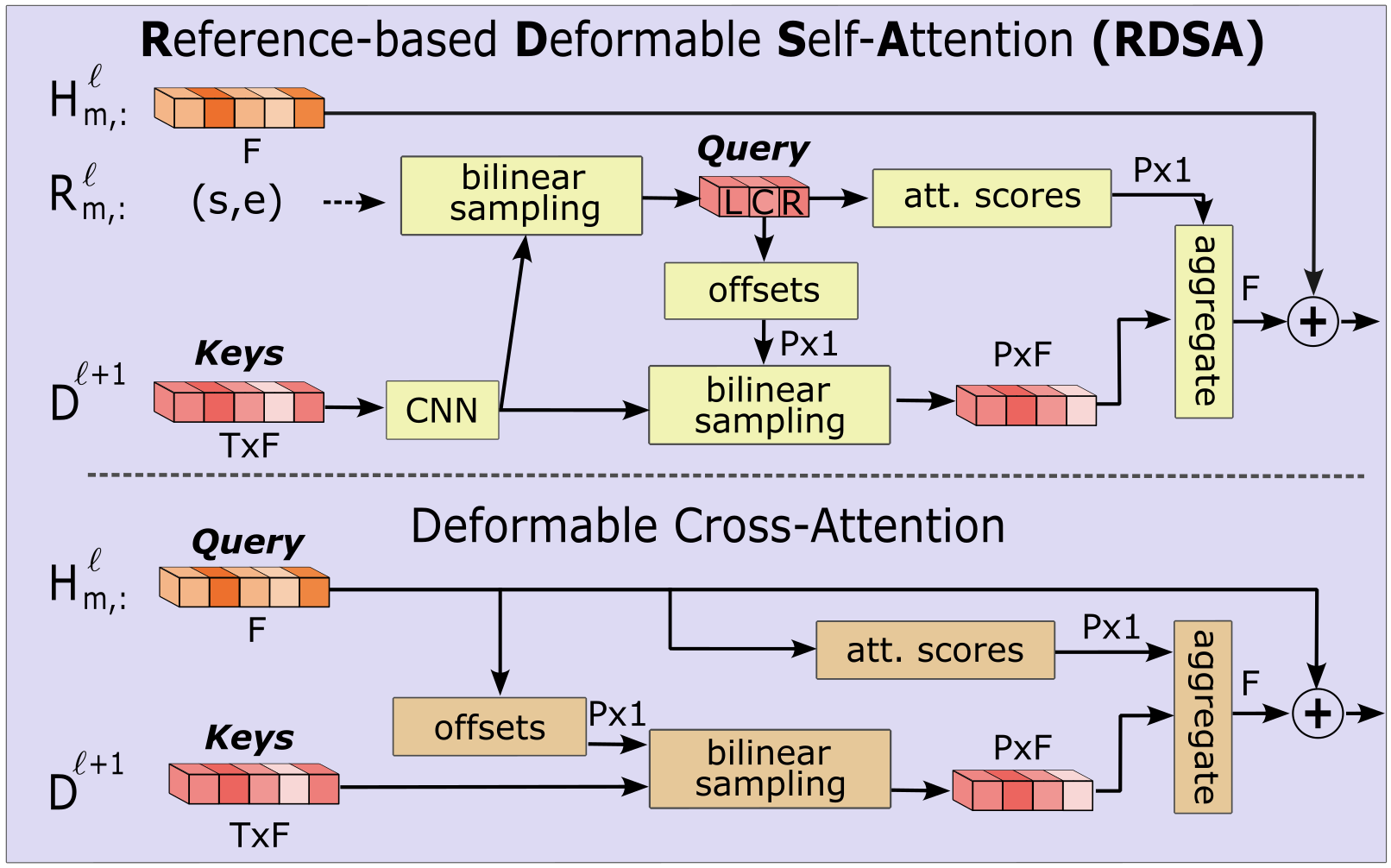}
        \vspace{-0.5cm}
        \caption{RDSA vs Def.CA\cite{xia2022vision} for a proposal $m$.}
        \label{fig:deformable_comparison}
    \end{subfigure}
    \vspace{-0.2cm}
    \caption{Our method (left) first processes the video and textual inputs using \cite{wang2025internvideo2}, and then recursively applies a shared SDST, a recurrent dual-stream model attached to the last $K$ layers. This iteratively refines dense embeddings and learnable \textit{recurrent decoder queries}. The SDST consists of a dense stream $\mathcal{D}$ for temporal and multi-modal dense refinement, and a sparse stream $\mathcal{S}$ that refines the recurrent queries conditioned on the dense signal. $\mathcal{S}$ includes the RDSA module (right), a context enhanced deformable attention mechanism that restricts the selectable keys based on the center, left-, and right-most action embeddings. The final dense embedding solves HD, while the recurrent queries address MR.}
    
    \label{fig:combined_figure}
    \vspace{-0.5cm}
\end{figure*}

\subsection{Problem definition}
In this paper, we address the problem of MR and HD based on textual descriptions. For this, we consider an arbitrary input-query pair $(\mathbf{X}^v, \mathbf{X}^t)$ where $\mathbf{X}^v\in \mathbb{R}^{T\times H \times W \times 3}$ and $\mathbf{X}^t \in \mathbb{R}^{L \times F_e}$. Here $T$ and $L$ are the number of frames and tokens, and $F_e$ the textual encoding dimension. Our goal is to predict, for the given video-query pair, its respective frame-wise saliency scores $\mathbf{Y}^{s} \in \mathbb{R}^{T}$ for HD, and the $M$ action moments $\mathbf{Y}^{m} \in \mathbb{R}^{M \times 2}$ for MR.

\subsection{Overview}
In our work we propose a dual-stream ST architecture that includes a dense (frame-level) and a sparse (segment-level) stream for MR and HD respectively (see Fig.~\ref{fig:main_architecture}). More specifically, we first leverage a frozen InternVideo2-1B~\cite{wang2025internvideo2} backbone, a model that possesses powerful spatio-temporal modeling capabilities, to extract $K$ intermediate visual-textual representations. These are then processed by multiple weight-sharing SDST layers, which refine dense representations and recurrent decoder queries. More in detail, at each level, the model refines the dense embeddings via multi-modal textual conditioning as well as temporal modeling. This signal is then used to condition the sparse stream $\mathcal{S}$, that builds on \cite{li2022dn} while incorporating our novel deformable attention mechanism, namely \textit{Reference-based Deformable Self-Attention} (see Fig.~\ref{fig:deformable_comparison}). This mitigates the shortcomings of the deformable CA mechanism, enhancing the contextual information of the decoder queries and thus improving the quality of the selected keys. After unrolling this process $K$ times, we apply the dense and the sparse prediction heads to compute the saliency scores and the predicted segment boundaries, respectively. 

\subsection{Sparse-Dense Side-Tuner (SDST)}

\subsubsection{Recurrent refinement with side-tuners}
\vspace{-0.2cm}
Our work builds on an ST framework, which first requires extracting $K$ intermediate video and textual features $\Tilde{\textbf{V}} \in \mathbb{R}^{K \times T \times F_v}$ and $\Tilde{\textbf{T}} \in \mathbb{R}^{K \times L \times F_t}$ (find the details in Sec.~\ref{sec:extraction_intermediate_reps}). Here $F_v$ and $F_t$ are their respective dimensionalities.  We then define a zero-initialized dense embedding $\mathbf{D}^0\in \mathbb{R}^{T \times F}$ and a set of $M$ learnable recurrent moment proposals inspired by \cite{liu2022dab}, which we refer to as \textit{recurrent decoder queries}. The queries include the learnable center-width moment references $\mathbf{R}^0 \in \mathbb{R}^{M \times 2}$, and their corresponding latent embeddings $\mathbf{H}^0 \in \mathbb{R}^{M \times F}$. We then define the recurrence for a given level $1\le \ell \le K$ as follows:
\begin{align}
\small
    \mathbf{D}^{\ell +1}, \mathbf{R}^{\ell +1}, \mathbf{H}^{\ell +1} = SDST(\mathbf{D}^\ell, \mathbf{R}^\ell, \mathbf{H}^\ell, \Tilde{\mathbf{V}}^\ell, \Tilde{\mathbf{T}}^\ell),
\label{eq:input}
\end{align}
where SDST represents the shared-weight layers of our model, progressively refining the representations across two streams: the \textit{1) dense learning stream} $\mathcal{D}$ and the \textit{2) sparse learning stream} $\mathcal{S}$, that we describe below. 
\vspace{-0.2cm}
\subsubsection{Dense learning stream $\mathcal{D}$}
\vspace{-0.2cm}
Following \cite{liu2024r}, this stream first refines the dense embeddings $\mathbf{D}^\ell$ by conditioning it to the video-textual embeddings $\Tilde{\mathbf{V}}^\ell$ and $\Tilde{\mathbf{T}}^\ell$. For this, both of these multi-modal embeddings are projected to a shared $F$-dimensional space, using two MLPs ($\mathcal{F}_v:F_v \rightarrow F$ and $\mathcal{F}_t: F_t \rightarrow F$):
\begin{align}
\small
\label{eq:raw_reps_after_proj}
    \mathbf{V}^\ell = \mathcal{F}_v(\Tilde{\mathbf{V}}^\ell)\;,\; \mathbf{T}^\ell = \mathcal{F}_t(\Tilde{\mathbf{T}}^\ell).
\end{align}
Then we incorporate the clip-wise visual information to $\mathbf{D}^\ell$ via a weighted sum modulated by $\beta^\ell\in [0,1]$, a zero-initialized layer-dependent parameter that yields:
\begin{align}
    \small
    \mathbf{D}^\ell := \beta^\ell \mathbf{D}^\ell + (1-\beta^\ell) \mathbf{V}^\ell.
\end{align}
Thereafter, we inject the textual information with CA, where $\mathbf{T}^\ell$ does not include the CLS token. We also apply a temporal modeling module defined as a SA and a subsequent point-wise feed forward network (PFFN):
\begin{align}
    \small
    \mathbf{D}^{\ell +1} = PFFN(SA(CA(\mathbf{D}^\ell, \mathbf{T}^\ell, \mathbf{T}^\ell))),
\end{align}

\vspace{-0.3cm}
\subsubsection{Sparse learning stream $\mathcal{S}$}
This stream can be seen as a recurrent DETR-based mechanism that exploits the complementarity of both sparse (MR) and dense (HD) tasks to refine the recurrent decoder queries. This is, refining the center-width references $\mathbf{R}^\ell\in \mathbb{R}^{M \times 2}$ and their latent embeddings $\mathbf{H}^\ell\in \mathbb{R}^{M \times F}$. Similar to the stream $\mathcal{D}$, we first incorporate a CA module that conditions $\mathbf{H}^\ell$ to the textual query. This is followed by a SA module to enable the information flow between different moment proposals:
\begin{align}
    \small
    \label{eq:ca_sa_injection}
    \mathbf{H}^\ell = SA(CA(\mathbf{H}^\ell, \mathbf{T}^\ell, \mathbf{T}^\ell)).
\end{align}
Another essential aspect of this stream is effectively conditioning the recurrent decoder queries to the video modality. Existing works~\cite{liu2022dab, li2022dn, gordeev2024saliency} typically rely on the use of deformable attention due to its convergence and performance benefits~\cite{zhu2020deformable, xia2022vision}. This mechanism leverages attention queries to predict offsets that define a limited set of selectable keys. However, in this task we observe that 1) these offsets collapse near their initialization and 2) are unable to \textit{look} beyond the current estimated boundaries (see Sec.~\ref{sec:ablation_deformable_attention_main_text}). We attribute this to contextual limitations, which motivates our proposed alternative \textit{Reference-based Deformable Self-Attention} (RDSA), enabling an enhanced cross-modality injection (see Sec.~\ref{sec:rdsa_main_text}). We also incorporate a simple PFFN to enhance the model's expressivity:
\begin{align}
    \small
    \mathbf{H}^\ell = PFFN(RDSA(\mathbf{R}^\ell, \mathbf{H}^\ell, \mathbf{D}^{\ell +1})).
\end{align}
Importantly, all of these modules are shared across the $K$ levels for parameter efficiency, and include residual connections and DropPath~\cite{larsson2016fractalnet} for stability and regularization.
\vspace{-0.3cm}
\subsubsection{Reference-based deformable self-attention}\label{sec:rdsa_main_text}
\noindent \textbf{Deformable attention in the context of existing anchor-free methods:}
One of the keys to the success of recent anchor-free methods like DETR --including works tackling VTG~\cite{gordeev2024saliency}-- is the use of deformable attention mechanisms~\cite{zhu2020deformable} for cross-modal interaction. This mechanism is known to provide faster convergence speed, reduced time complexity, and as shown in Sec.~\ref{sec:supp_ablation_deformable_att}, overall better performance than standard CA. As depicted in Fig.~\ref{fig:deformable_comparison} (down), the key of this module is the lack of explicit interaction between queries and keys, which are defined as
\begin{align}
    \small
    \mathbf{Q}=\mathbf{X}_\mathcal{Q}\mathbf{W}_\mathcal{Q}^{def} \;,\; \mathbf{K}=\mathbf{X}_\mathcal{K}\mathbf{W}_\mathcal{K}^{def},
    \label{eq:def_att_projections}
\end{align}
where $\mathbf{W}_\mathcal{Q}^{def}$, $\mathbf{W}_\mathcal{K}^{def}$ are two linear projections. The deformable attention thus avoids computing the $\textbf{QK}^T$ similarity matrix, and instead employs an offset and attention-score predictors, $\mathcal{G}_\Delta$ and $\mathcal{G}_\textbf{A}$ to select and weight --only based on the queries-- a small subset $P$ of selectable keys. The output of this mechanism, the weighted aggregation of the selected keys, namely $\textbf{S}$, is computed as follows
\begin{align}
    \label{eq:standard_offset_att_prediction}
    \small
    \Delta= \mathcal{G}_\Delta(\mathbf{Q}) \in \mathbb{R}^{M \times P}, \; \mathbf{A} = \mathcal{G}_{\textbf{A}}(\mathbf{Q}) \in \mathbb{R}^{M \times P}, \\
    \mathbf{S} = \sum_{p=1}^P \left(\mathbf{A}_{:,p} \; \mathbf{K}[{\textbf{c} + \textbf{w} \odot \Delta_{:,p}}] \right) \in \mathbb{R}^{M\times F}
    \label{eq:deformable_att}
\end{align}
where $\mathcal{G}_\Delta$, $\mathcal{G}_\textbf{A}$ are often modeled as CNNs and $x[y]$ is the bilinear sampling of $x$ on index(es) $y$. Notably, in our 2-dimensional setup, the offsets $\Delta_{:,p}$ are added to center reference $\textbf{c}=\mathbf{R}_{:,0}^\ell \in \mathbb{R}^{M}$, and weighted over the action width $\textbf{w}=\mathbf{R}_{:,1}^\ell \in \mathbb{R}^{M}$. See Sec.~\ref{sec:supp_ablation_deformable_att} for additional details.

\noindent\textbf{Limitations:} Interestingly, in the literature we find that when it comes to the deformable attention, most works operate with a certain independence to whether they deal with a deformable self-attention or cross-attention, assuming that the benefits of this mechanism extrapolate to both scenarios. In this work, however, we find that in fact, the deformable attention is only naturally suited for \underline{self-attention} scenarios as its effectiveness inherently relies on \textit{implicit interactions} between $\mathbf{Q}$ and $\mathbf{K}$ --not present in the cross-attention case. Concretely, in a self-attention setup, $\mathbf{X}_\mathcal{Q} = \mathbf{X}_\mathcal{K}$ (see Eq.~\ref{eq:def_att_projections}). Thus, modeling $\mathcal{G}_{\Delta}$ and $\mathcal{G}_{\textbf{A}}$ as CNNs naturally allows queries to gain context of the local neighborhood of queries, and consequently, of the key/value space. This breaks their independence assumption, giving the queries critical information to decide \textit{where to look}. 

Importantly, this is not the case with the deformable \underline{cross attention} where we try to inject cross-modal information from the keys to the queries, which is our goal in this this work. In this scenario, $\mathbf{X}_\mathcal{Q} \neq \mathbf{X}_\mathcal{K}$, and thus, using CNN-based offset predictors does not inject knowledge of the key/value space, making its guesses \textit{blind} or based on the \textit{blind} predictions of previous iterations. Concretely, in Sec.~\ref{sec:ablation_deformable_attention_main_text} we show that in our task, the drawbacks of this naive deformable cross-attention have various empirical implications such as the offset collapse near the offset initialization values, or the inability to select keys from frames beyond the estimated action boundaries, critical to refine segments into longer actions. This motivates our proposed \textit{Reference-based Deformable Self-Attention}, an alternative mechanism that allows reformulating the deformable CA module into a deformable SA based on the references $\textbf{R}^\ell$.

\noindent \textbf{Our proposed alternative (RDSA):} Our proposed RDSA (see Fig.~\ref{fig:deformable_comparison}) takes as input the center-width references $\mathbf{R}^\ell \in \mathbb{R}^{M \times 2}$ and the dense embeddings $\mathbf{D}^\ell\in \mathbb{R}^{T \times F}$. Then, differently from other methods that leverage the learnable queries $\textbf{Q}=\textbf{H}^\ell$ to compute the offsets and attention scores (see Eq.~\ref{eq:standard_offset_att_prediction}), we propose using an alternative query embedding. Concretely, we first refine the dense embedding $\mathbf{D}_l$ with a simple CNN, allowing frames to gain local context of their surroundings. Then, we use bilinear sampling to extract three key action embeddings: left-most (l), center (c), and right-most (r). The center provides highly informative action features, while the extremes help refine the moment boundaries. This yields,
\begin{align}
\small
\label{eq:rdsa_new_queries}
\mathbf{\hat{Q}} = \mathbf{\hat{X}}_\mathcal{Q} \mathbf{W}_\mathcal{Q}^{def} \;,\; \mathbf{\hat{X}}_\mathcal{Q} = CNN(\mathbf{D}^{\ell})[l,c,r].
\end{align}
where $\mathbf{\hat{X}}_\mathcal{Q} \in \mathbb{R}^{M \times 3F}$. This rewrites Eq.~\ref{eq:standard_offset_att_prediction} to
\begin{align}
\small
\hat{\Delta} = \mathcal{G}_\Delta(\mathbf{\hat{Q}}), \; \hat{\mathbf{A}} = \mathcal{G}_\textbf{A}(\mathbf{\hat{Q}}).
\end{align}
Finally, we apply the deformable attention from Eq.~\ref{eq:deformable_att} using the new offsets and attention scores $\hat{\Delta}$ and $\mathbf{\hat{A}}$. This can be viewed as \textit{asking the key spots of the action where to look} based on the textual information that we previously injected in Eq.~\ref{eq:ca_sa_injection}. This naturally solves the context limitation of the standard deformable CA given that $\mathbf{\hat{X}}_\mathcal{Q}$ and $\mathbf{X}_\mathcal{K}$ now live in the same latent space --as they are both derived from $\textbf{D}^\ell$. This allows the model to make informed decisions about which keys to attend to while keeping all its core computational advantages. 

\subsection{Prediction heads and training objectives}\label{sec:prediction_head_and_training_objectives}
After recursively applying the SDST over $K$ intermediate layers, we compute the respective saliency scores and action segment proposals, as well as define their corresponding objective functions. We briefly describe these below, but refer interested readers to Sec.~\ref{sec:supp_objective_functions} for more details. 

\noindent \textbf{Highlight Detection:} To solve HD, we define the saliency scores as the cosine similarity between the dense embedding $\mathbf{D}^K$ and a pooled representation of $ \mathbf{T}^K$. We then apply an InfoNCE loss~\cite{oord2018representation} that learns a ranking of these scores
\begin{align}
    \small
    \mathcal{L}_{HD} = \lambda_{0} \mathcal{L}_{InfoNCE}.
\end{align}
\noindent \textbf{Moment Retrieval:} To solve MR we first apply the Hungarian algorithm to obtain a one-to-one matching between the predicted queries and the GT actions. We then project $\mathbf{H}^\ell$ to compute the action probability $\textbf{p}\in \mathbb{R}^{M}$ of the $M$ different queries, which we train with a FocalLoss~\cite{lin2017focal}. We also compute the moment boundaries $\mathbf{\hat{Y}}^m \in \mathbb{R}^{M \times 2}$ which we optimize using an L1 and an IoU loss. Finally, we use an L1 loss to learn an actionness score, which predicts the maximum IoU of each query w.r.t. the GT actions
\begin{align}
    \small
    \mathcal{L}_{MR} =  \lambda_{1} \mathcal{L}_{act} + \sum_{l=1}^\ell \lambda_{2} \mathcal{L}_{cls}^\ell + \lambda_{3} \mathcal{L}_{l1}^\ell + \lambda_{4} \mathcal{L}_{IoU}^\ell.
\end{align}
Note that all but the $\mathcal{L}_{act}$ term are optimized across different refinement levels to promote a faster convergence.

\noindent \textbf{Alignment losses:} To improve the video-textual alignment, we define $\mathcal{L}_{align}$ based on two SampledNCE losses~\cite{yang2020understanding} to align $\mathbf{V}$ and $\mathbf{T}$ embeddings across the different refinement levels. We apply this along both the batch and the intermediate-layer dimensions. This pulls the action embeddings closer to the textual representations while learning complementary information across intermediate layers.

\noindent \textbf{Final loss:} The final loss is defined as
\begin{align}
    \small
    \mathcal{L} = \lambda_{5} \mathcal{L}_{HD} + \lambda_{6} \mathcal{L}_{MR} + \lambda_{7}\mathcal{L}_{align}.
\end{align}

%\noindent \textbf{Inference:} During inference we apply an NMS post-processing to filter-out action predictions. This algorithm sorts them according to the action confidence scores, defined as the squared root of the product of the class probabilities and actionness scores. 

\vspace{-0.3cm}
\section{Extracting intermediate representations of InternVideo2} \label{sec:extraction_intermediate_reps}
The cornerstone of ST is leveraging intermediate multi-modal representations from frozen VLMs. While previous ST approaches predominantly rely on CLIP, we argue that CLIP has notable limitations in temporal modeling and aligning visual features with textual queries beyond simple static descriptions. To overcome these limitations, we incorporate InternVideo2~\cite{wang2025internvideo2} into our framework, leveraging its advanced spatio-temporal modeling capabilities~\cite{gordeev2024saliency}. However, extracting intermediate representations from InternVideo2 --particularly from its visual encoder $\mathcal{E}_v$-- remains a challenging and underexplored task in the literature, despite its significant impact on performance.

In particular, the main challenge arises from how to pool the spatio-temporal token dimension $L_v$ of $\hat{\textbf{V}}^\ell \in \mathbb{R}^{T \times L_v \times F}$, the output embedding of the $\ell$-th layer of $\mathcal{E}_v$, into the final intermediate features $\Tilde{\textbf{V}}^{\ell} \in \mathbb{R}^{T \times F}$ (see Eq.~\ref{eq:input}). Prior CLIP-based methods simply use the CLS token as "summaries" over $L_v$. However, we find this strategy to be suboptimal for InternVideo2 (see Sec.~\ref{sec:ablation_internvideo2_feats}) as this limits its spatial-aggregation capabilities. Notably, \cite{wang2025internvideo2} optimizes an AdaptivePool module that computes the final-layer embeddings, which are then aligned with the text. We conjecture that leveraging this enhanced multi-modal alignment is key to performance. Unfortunately, such pooled representation is only computed at the last layer, leaving the question of \textit{how to pool the remaining intermediate layers.} Ideally, we would optimize layer-independent AdaptivePool modules, but this would require full back-propagation through the entire model. A more efficient alternative would be computing gradients of the AdaptivePool modules only. This is, however, still computationally infeasible, not for the gradients themselves, but because of the need to load $L_v$ spatio-temporal tokens per frame. This increase in the input memory size translates, for instance, to a $15\times$ memory increase for QVHighlights. We thus hypothesize that re-using this frozen pooling module across the $K$ intermediate layers allows a memory efficient training --no additional backpropagation nor memory requirements are needed-- while leveraging its enhanced pooling capabilities, even despite their respective distribution shifts. Formally, we compute
\begin{align}
    \small
    \label{eq:adaptive_pool_internvideo2}
    \Tilde{\textbf{V}}^{\ell} = AdaptivePool\left(\hat{\textbf{V}}^{\ell}\right) \in \mathbb{R}^{T \times F},\; 1 \leq \ell \leq K.
\end{align}
As shown in Sec.~\ref{sec:ablation_internvideo2_feats}, this pooling strategy considerably improves other alternatives, demonstrating the benefits of this module re-utilization.

\vspace{-0.1cm}
\section{Experimentation}

\subsection{Experimental setup}
We test our proposal on various datasets for MR and HD, namely QVHighlights~\cite{lei2021detecting}, TACoS~\cite{rohrbach2014coherent}, and the Charades-STA~\cite{gao2017tall} dataset. Further details of the datasets can be found in Sec.~\ref{sec:extended_experimental_setup}. For the task of MR on QVHighlights, we compute Recall@1 with two IoU thresholds, 0.5 and 0.7, and the mean average precision (mAP) at IoU thresholds between 0.5 and 0.95 with a step of 0.05 --i.e., [0.5:0.05:0.95]. For HD, we report mAP and HIT@1 over the positive frames --i.e., the most salient ones classified as "VeryGood". For TACoS and Charades-STA, we compute Recall@1 at 0.3, 0.5, and 0.7 IoU thresholds, as well as mIoU.

\subsection{Main experimental results}\label{sec:main_experimental_results}

We begin by comparing the performance of SDST over various relevant baselines. Concretely, observe in Tab.~\ref{tab:main_results_qvhighlights} the evaluation of our method on the \textit{test} and \textit{val} splits of QVHighlights. These results indicate that our method considerably outperforms R$^2$-Tuning. Similarly, SDST performs very competitively and even surpasses on several metrics the current SOTA method, namely SG-DETR~\cite{gordeev2024saliency}. Note however that our method attains this performance while having only a $27\%$ of the parameter count of SG-DETR. We refer interested readers to Sec.~\ref{sec:supp_statistical_significance} for the statistical significance analysis.

% Table main results QVHighglights
\begin{table*}
\footnotesize
\centering
\resizebox{0.88\textwidth}{!}{
\begin{tabular}{c|ccccc|cc|ccccc|cc|c }
\toprule

\multirow{4}{*}{\textbf{Method}} & \multicolumn{7}{c|}{\textbf{test}} & \multicolumn{7}{c|}{\textbf{val}} & \multirow{4}{*}{\textbf{\#Params}} \\

\cmidrule(lr){2-8}
\cmidrule(lr){9-15}

& \multicolumn{5}{c|}{\textbf{MR}} & \multicolumn{2}{c|}{\textbf{HD}} & \multicolumn{5}{c|}{\textbf{MR}} & \multicolumn{2}{c|}{\textbf{HD}} &  \\

\cmidrule(lr){2-8}
\cmidrule(lr){9-15}

 & \multicolumn{2}{c}{\textbf{R1}} & \multicolumn{3}{c|}{\textbf{mAP}} & \multicolumn{2}{c|}{\textbf{$\ge$ Very good}} & \multicolumn{2}{c}{\textbf{R1}} & \multicolumn{3}{c|}{\textbf{mAP}} & \multicolumn{2}{c|}{\textbf{$\ge$ Very good}} & \\
 
 & \textbf{@0.5}  & \textbf{@0.7} & \textbf{@0.5} & \textbf{@0.75} & \textbf{Avg.} & \textbf{mAP} & \textbf{HIT@1} & \textbf{@0.5}  & \textbf{@0.7} & \textbf{@0.5} & \textbf{@0.75} & \textbf{Avg.} & \textbf{mAP} & \textbf{HIT@1} & \\
\midrule
\rowcolor{lightgray} BeautyThumb & -- & -- & -- & -- & -- & 14.36 & 20.88 & -- & -- & -- & -- & -- & -- & -- & --\\

\rowcolor{lightgray} DVSE & -- & -- & -- & -- & -- & 18.75 & 21.79 & -- & -- & -- & -- & -- & -- & -- & -- \\

\rowcolor{lightgray} MCN & 11.41 & 2.71 & 24.94 & 8.22 & 10.67  & -- & -- & -- & -- & -- & -- & -- & -- & -- &--  \\
\rowcolor{lightgray} CAL & 25.49 & 11.54 & 23.40 & 7.65 & 9.89 & -- & -- & -- & -- & -- & -- & -- & -- & -- & --  \\
XML+ & 46.69 & 33.46 & 47.89 & 34.67 & 34.90 & 35.38 & 55.06 & -- & -- & -- & -- & -- & -- & -- & -- \\
Moment-DETR & 52.89 & 33.02 & 54.82 & 29.40 & 30.73 & 35.69 & 55.60 & 53.94 & 34.84 & -- & -- & 32.20 & 35.65 & 35.65 & 4.8M \\
UMT & 56.23 & 41.18 & 53.83 & 37.01 & 36.12 & 38.18 & 59.99 & 60.26 & 44.26 & 56.70 & 39.90 & 38.59 & 39.90 & 64.20 & 14.9M  \\
\rowcolor{lightgray} MomentDiff & 58.21 & 41.48 & 54.57 & 37.21 & 36.84 & -- & -- & -- & -- & -- & -- & -- & -- & -- & -- \\
QD-DETR & 62.40 & 44.98 & 62.52 & 39.88 & 39.86 & 38.94 & 62.40 & 62.68 & 46.66 & 62.23 & 41.82 & 41.22 & 39.13 & 63.03 & 7.6 M \\
MH-DETR & 60.05 & 42.48 & 60.75 & 38.13 & 38.38 & 38.22 & 60.51 & 60.84 & 44.90 & 60.76 & 39.64 & 39.26 & 38.77 & 61.74 & 8.2 M \\
UniVTG & 58.86 & 40.86 & 57.60 & 35.59 & 35.47 & 38.20 & 60.96 & 59.74 & -- & -- & -- & 36.13 & 38.80 & 61.8 & 41.3M \\
TR-DETR & 64.66 & 48.96 & 63.98 & 43.73 & 42.62 & 39.91 & 63.42 & 67.10 & 51.48 & 66.27 & 46.42 & 45.09 & -- & -- & 7.9 M\\
CG-DETR & 65.43 & 48.38 & 64.51 & 42.77 & 42.86 & 40.33 & 66.21 & 67.35 & 52.06 & 65.57 & 45.73 & 44.93 & 40.80 & 66.70 & 12.0 M\\

\rowcolor{lightgray} 
BAM-DETR & 62.71 & 48.64 & 64.57 & 46.33 & 45.36 & -- & -- & 65.10 & 51.61 & 65.41 & 48.56 & 47.61 & -- & -- & -- \\

EaTR & -- & -- & -- & -- & -- & -- & -- & 61.36 & 45.79 & 61.86 & 41.91 & 41.74 & 37.15 & 58.65 & 9.0 M\\
\rowcolor{lightgray} 
Mr. BLIP & 74.77 & 60.51 & 68.12 & 53.38 & -- & -- & -- & 76.13 & 63.35 & 69.39 & 55.78 & -- & -- & -- & 19.0 M \\
\rowcolor{lightgray} 
LLaVA-MR & 76.59 & 61.48 & 69.41 & 54.40 & -- & -- & -- & 78.13 & 64.13 & 69.64 & 56.32 & -- & -- & -- & 17.0 M \\
\rowcolor{lightgray} 
HL-CLIP & -- & -- & -- & -- & -- & 41.94 & 70.60 & -- & -- & -- & -- & -- & 42.37 & 72.40 & 2.0 M \\
R\textsuperscript{2}-Tuning & 68.03 & 49.35 & 69.04 & 47.56 & 46.17 & 40.75 & 64.20 & 68.71 & 52.06 & -- & -- & 47.59 & 40.59 & 64.32 & 2.7 M \\
SG-DETR\textsuperscript{\textdagger} & \textbf{72.20} & \textbf{56.60} & \textbf{73.20} & \textbf{55.80} & \textbf{54.10} & \textbf{43.76} & \textbf{69.13} & -- & -- & \textbf{73.52} & \textbf{57.91} & \textbf{55.64} & \underline{43.91} & \underline{71.47} & 15.0 M \\
Flash-VTG\textsuperscript{\textdagger} & 70.69 & 53.96 & \underline{72.33} & 53.85 & 52.00 & -- & -- & 73.10 & 57.29 & 72.75 & 54.33 & 52.84 & -- & -- & 10.9 M \\
\midrule
\textbf{Ours}\textsuperscript{\textdagger} & \underline{70.82} & \underline{56.23} & 71.31 & \underline{54.99} & \underline{53.31} & 43.40 & \textbf{69.13} & \textbf{73.68} & \textbf{60.90} & \textbf{73.52} & \underline{57.42} & \underline{55.60} & \textbf{44.00} & \textbf{72.00} & 4.1 M \\
\bottomrule
\end{tabular}}
\vspace{-0.2cm}
\caption{MR and HD results for QVHighlights \textit{test} and \textit{val} split.{}\textsuperscript{\textdagger} indicates the use of InternVideo2 features. \textbf{Bold} stands for the best and \underline{underline} for the second-best, considering only works supporting both MR and HD. The rest are marked in gray.} \label{tab:main_results_qvhighlights}
\vspace{-0.5cm}
\end{table*}

Most of the presented methods in Tab.~\ref{tab:main_results_qvhighlights} rely on alternative backbones different from InternVideo2 --e.g., CLIP or Slowfast. To establish a more fair comparison, in Fig.~\ref{fig:bubble_plot} we follow the work of \cite{gordeev2024saliency} and evaluate a set of relevant baselines when using InternVideo2 features --find the complete ablation in Sec.\ref{sec:supp_ablation_fair_comparison}. Observe that our work performs on par with the SG-DETR and even more importantly to our work, the SDST very considerably outperforms R$^2$-Tuning, improving the MR performance by a $3.82\%$ average mAP or the HD by a $2.21\%$ mAP. This empirically demonstrates the benefits of leveraging a proposal-free architecture over a more rigid anchor-based method. Also find in Sec.~\ref{sec:comparison_other_petl_metl_methods} an extended comparison with other ST works.

To complement our findings, in Tab.~\ref{table:main_results_charades_tacos} we evaluate our method on the Charades-STA and TACoS benchmarks. Observe that our method obtains SOTA in both of these datasets. In Charades-STA, for instance, improving by $2.71\%$ of R1@0.7 and $2.06\%$ of mIoU, the existing SOTA. Similarly, in TACoS, SDST improves SG-DETR by $2.39\%$ R1@0.7 and $1.27\%$ mIoU.

% Main results Charades-STA and TACoS
\begin{table}
\centering
\resizebox{0.36\textwidth}{!}{
\begin{tabular}{l|ccc|ccc}
\toprule
\textbf{Method} & \multicolumn{3}{c}{\textbf{Charades-ST}} & \multicolumn{3}{c}{\textbf{TACoS}} \\
& R@0.5 & R@0.7 & mIoU &  R@0.5 & R@0.7 & mIoU \\
\midrule
M-DETR & 53.6 & 31.4 & - & 24.7 & 12.0 & 25.5 \\
UMT & 48.3 & 29.3 & - & - & - & - \\
UniVTG & 58.0 & 35.7 & 50.1 & 35.0 & 17.4 & 33.6 \\
QD-DETR & 57.3 & 32.6 & - & 36.8 & 21.1 & 35.8 \\
CG-DETR & 58.4 & 36.3 & 50.1 & 39.6 & 22.2 & 36.5 \\
BAM-DETR & 60.0 & 39.4 & 52.3 & 41.5 & 26.8 & 39.3 \\
TR-DETR & 57.6 & 33.5 & - & - & - & - \\
MR. BLIP & 69.3 & 49.3 & 58.6  & - & - & - \\
LLaVA-MR & \underline{70.6} & 49.6 & \underline{59.8} & - & - & - \\
R\textsuperscript{2}-Tuning & 59.8  & 37.0 & 50.9 & 38.7 & 25.1 & 35.9  \\
SG-DETR\textsuperscript{\textdagger} & 70.2 & 49.5 & 59.1 & \textbf{44.7} & \underline{29.9} & \underline{40.9} \\
FlashVTG\textsuperscript{\textdagger} & 70.3 & \underline{49.9} & -- & 41.8 & 24.7 & 37.6 \\

\midrule
\textbf{Ours}\textsuperscript{\textdagger} & \textbf{72.0} & \textbf{52.6} & \textbf{61.2} & \underline{44.5} & \textbf{32.3} & \textbf{42.2} \\
\bottomrule
\end{tabular}}
\vspace{-0.2cm}
\caption{Comparison on Charades-STA and TACoS datasets. {}\textsuperscript{\textdagger} indicates the use of InternVideo2 features. \textbf{Bold} stands for the best and \underline{underline} for the second-best.}
\vspace{-0.2cm}
\label{table:main_results_charades_tacos}
\end{table}

\vspace{-0.1cm}

\section{Ablation studies}
In this section, we ablate over various relevant aspects of our main contributions. Note that unless states otherwise, we follow the literature and evaluate them on the \textit{val} split of QVHighlights. Also, we write write in \textbf{bold} the best results.

\subsection{Leveraging InternVideo2 features for ST}\label{sec:ablation_internvideo2_feats}

\noindent\textbf{1) Does pooling matter?} As argued in Sec.~\ref{sec:extraction_intermediate_reps}, one of the main challenges in using the InternVideo2 backbone for ST is pooling the spatio-temporal clip-wise intermediate representations. In Tab.~\ref{tab:ablation_pooling_strategies} we compare different pooling strategies. Concretely, we compare the standard CLS-pooling, average-pooling, and our proposed re-utilization of the frozen AdaptivePool (Sec.~\ref{sec:extraction_intermediate_reps}). Observe that CLS-pooling results in a considerable performance degradation of up to $5.07\%$ average mAP on MR or $7.74\%$ HIT@1 on HD w.r.t. the adaptive pooling strategy. The average pooling partially mitigates this degradation, but still obtains a considerably decreased performance. This confirms the significant impact of a carefully chosen pooling strategy.

\begin{table}[t]
\centering
\vspace{-0.2cm}
\resizebox{0.45\textwidth}{!}{
\begin{tabular}{l|ccccc|cc}
\toprule
\textbf{Pool strat.} & \multicolumn{5}{c}{\textbf{MR}} & \multicolumn{2}{c}{\textbf{HD}} \\

& \textbf{R1@0.5} & \textbf{R1@0.7} & \textbf{mAP@0.5} & \textbf{mAP@0.75}& \textbf{mAP} & \textbf{mAP} & \textbf{HIT@1} \\
\midrule
CLS                & 66.45 & 52.65 & 67.96 & 51.32 & 50.53 & 41.01 & 64.26 \\
Avg. pool          & 70.65 & 56.9 & 70.43 & 54.58 & 53.44 & 43.06 & 69.68 \\
\midrule
Adapt. pool        & \textbf{73.68} & \textbf{60.90} & \textbf{73.52} & \textbf{57.42} & \textbf{55.60} & \textbf{44.00} & \textbf{72.00}  \\
\bottomrule
\end{tabular}}
\vspace{-0.2cm}
\caption{Effect of using different pooling strategies. }
\vspace{-0.5cm}
\label{tab:ablation_pooling_strategies}
\end{table}

\begin{figure}[t]
\centering
\includegraphics[width=0.48\textwidth]{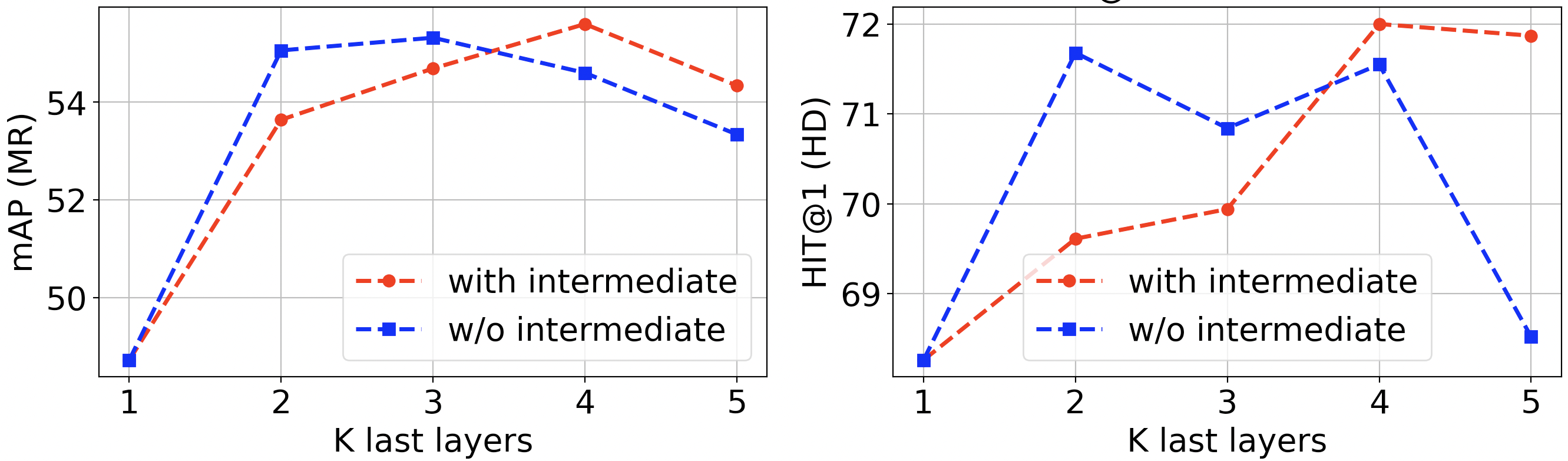}
\vspace{-0.6cm}
\caption{\label{fig:ablation_interm_feats} Ablation of the number of refinement levels and of their use of intermediate or last-layer features only.}
\vspace{-0.4cm}
\end{figure}

\noindent\textbf{2) Feature refinement vs. feature sampling:} One of the most critical decisions of ST is determining the number of intermediate levels that should be used. Normally, doing so involves evaluating the performance when refining the last $K$ intermediate layers~\cite{liu2024r} . As shown in Fig.~\ref{fig:ablation_interm_feats} (red) this indicates that in our case, the best performing $K$ is $4$. The literature often suggests this argument to be compelling enough to claim that using intermediate representations is beneficial over using only last-layer features. In this work, however, we find it reasonable to wonder: \textit{Does this analysis depict the importance of intermediate features, or is it simply evidencing the need to do multiple refinement steps, regardless of the chosen features?} 

To shed light on this issue, in Fig.~\ref{fig:ablation_interm_feats} (blue) we evaluate the performance when doing $K$ different refinement steps, but importantly, we always use the \underline{last-layer features} only. Observe that for $K=2$ and $K=3$, actually, this improves the homologous experiments with intermediate features. Differently, for $K=4$ and $K=5$, the use of last-layer features harms the performance. These experiments thus suggest that proving the usefulness of intermediate features is not as straightforward as previously thought. In fact, these indicate that the advantages of using intermediate features arise only as we consider shallower layers.

% TODO: Run the ablations of 2/3 levels of refinements but with deeper-level features, to ensure wheather it is a matter of refinement capacity or really just that these very shallow features are useless.

\noindent\textbf{3) Why not sampling features from even shallower layers?} As shown in Tab. \ref{tab:ablation_different_sampling_strategies_main_text}, this is not necessarily helpful because of what we call, \textit{depth-pooling trade-off}. We conjecture that sampling from shallower layers does in fact provide additional complementary information. Nevertheless, this inevitably results in a distribution shift w.r.t. the last-layer feature distribution, hindering the effectiveness of the frozen AdaptivePool (see Eq. \ref{eq:adaptive_pool_internvideo2}). The infeasibility of retraining this module for computational reasons, thus, leaves us with a delicate trade-off between the quality of the pooling and the depth of the features that must be accounted for.

\begin{table}
\centering
\resizebox{0.48\textwidth}{!}{
\begin{tabular}{l|ccccc|cc}
\toprule
\textbf{Levels} & \multicolumn{5}{c}{\textbf{MR}} & \multicolumn{2}{c}{\textbf{HD}} \\

& \textbf{R1@0.5} & \textbf{R1@0.7} & \textbf{mAP@0.5} & \textbf{mAP@0.75}& \textbf{mAP} & \textbf{mAP} & \textbf{HIT@1} \\
\midrule
$[37,38,39,40]$        & 73.68 & 60.90 & 73.52 & 57.42 & 55.60 & 44.0 & 72.00  \\
$[25,30,35,40]$        & 71.48 & 59.48 & 71.45 & 56.37 & 54.83 & 43.42 & 70.19 \\
$[10,20,30,40]$         & 69.03 & 55.35 & 69.56 & 53.37 & 52.47 & 42.73 & 68.39 \\
\bottomrule

\end{tabular}}
\vspace{-0.3cm}
\caption{Ablation of different sampling strategies of intermediate features. The first column indicates the $K=4$ sampled layers --of a total of 40 layers.}
\label{tab:ablation_different_sampling_strategies_main_text}
\vspace{-0.5cm}
\end{table}

\subsection{Study of deformable attention}\label{sec:ablation_deformable_attention_main_text}
%1) Basic comparison with other attention strategies. Maybe show the need for CNN to overcome the binarization issue. Also, we might need to argue it using the std to show that it is more data-dependent.

\noindent\textbf{1) Comparison w.r.t. other baselines:} Here we empirically evaluate the benefits of our proposed RDSA. Concretely, in Tab. \ref{tab:ablation_comparison_deformable_attention} we empirically compare the performance of the standard CA~\cite{vaswani2017attention} and Def.~CA~\cite{zhu2020deformable}, as well as different decoder-query initialization strategies~\cite{zhang2022dino} for the latter. Notice the prominent performance degradation derived from the use of the CA, or how the query initialization techniques --far from improving the performance of the Def.~CA-- in some cases are even harmful. In contrast, our method consistently outperforms all these tested baselines, improving by $2.71\%$ R1@0.7 or $1.62\%$ mAP w.r.t. the original Def.~CA. We refer to Sec.~\ref{sec:supp_ablation_deformable_att} for the ablation of the different sampling strategies of RDSA.

\begin{figure}[t]
\centering
\includegraphics[width=0.45\textwidth]{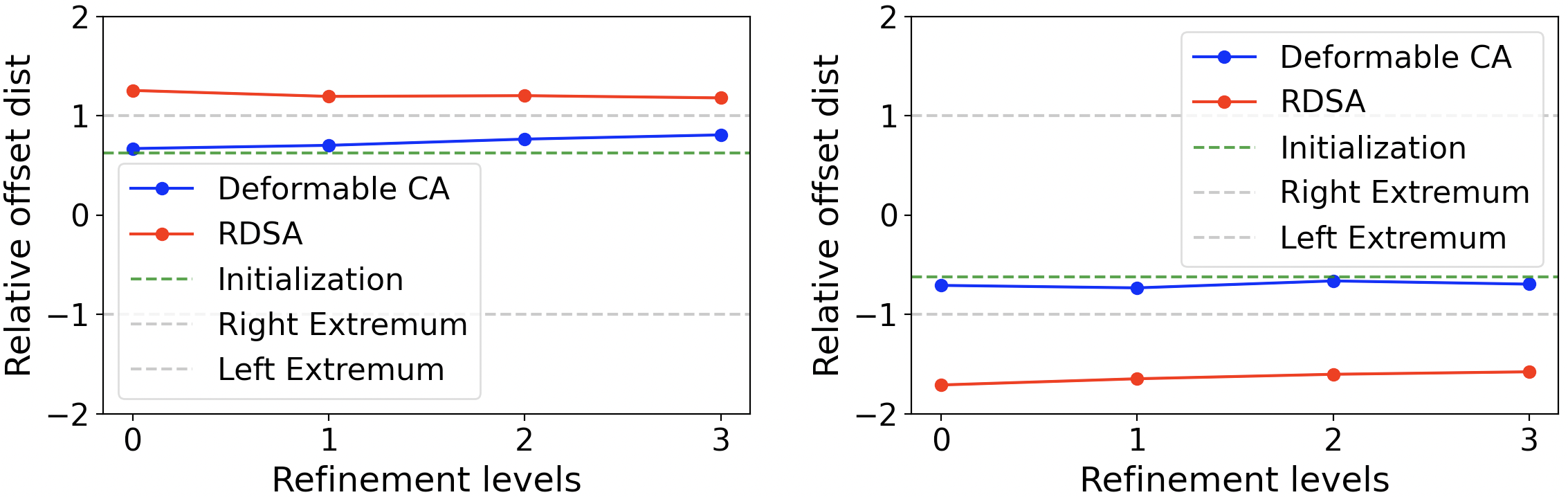}
\vspace{-0.3cm}
\caption{\label{fig:rdsa_comparison} Average of the weighted offsets across $M$ decoder queries and $N$ batch elements for the $K=4$ refinement levels. Here head 0 (left) is initialized near the left boundary, and head 1 (right) near the right boundary.}
\vspace{-0.5cm}
\end{figure}

\noindent \textbf{2) Where are the offsets pointing to?} In Fig.~\ref{fig:rdsa_comparison} we depict the refinement of the weighted offset of a given query $q$, defined as $d_{q} = \sum_{p=1}^P A_{q,p} \Delta_{q,p}$, across the $K$ different refinement steps. Note that as depicted in Eq.~\ref{eq:deformable_att}, these offsets are relative to the estimated moment widths, with -1 and +1 indicating the estimated left- and right-most boundaries. The Def.~CA lacks context awareness, causing its offset predictions to stay near the offset initializations. In contrast, RDSA leverages an enhanced contextualization, thus reducing its bias towards the offset initialization. Notably, RDSA learns to \textit{look} beyond the current action boundaries (offsets $<-1$ and $>+1$) capturing near-boundary information, crucial for refining moment boundaries. This contrasts with \cite{zhu2020deformable} which remains constrained within the predicted moment boundaries. We consider this the reason why RDSA is especially useful to localize long actions (see Sec.~\ref{supp:details_deformable_attention}).

\begin{table}
\centering
\resizebox{0.48\textwidth}{!}{
\begin{tabular}{l|ccccc|cc}
\toprule
\textbf{Att. strat.} & \multicolumn{5}{c}{\textbf{MR}} & \multicolumn{2}{c}{\textbf{HD}} \\

& \textbf{R1@0.5} & \textbf{R1@0.7} & \textbf{mAP@0.5} & \textbf{mAP@0.75} &\textbf{mAP} & \textbf{mAP} & \textbf{HIT@1} \\
\midrule
Stand. CA~\cite{vaswani2017attention}      & 70.77 & 51.74 & 66.78 & 45.19 & 42.72 & 43.16 & 69.87          \\

Def.~CA~\cite{zhu2020deformable}        & 72.58 & 58.19  & 71.82 & 55.80 & 54.27 & 43.26   & 70.58    \\

Def.~CA +PureInit~\cite{zhang2022dino} & 72.13 & 57.87 & 70.68 & 54.85 & 52.92 & 43.19 & 70.32         \\
Def.~CA +MixedInit~\cite{zhang2022dino} & 72.26 & 58.71 & 71.94 & 56.43 & 54.94 & 43.10 & 69.86          \\

\midrule

Ours & \textbf{73.68} & \textbf{60.90} & \textbf{73.53} & \textbf{57.42} & \textbf{55.60} & \textbf{44.0} & \textbf{72.00}         \\

\bottomrule

\end{tabular}}
\vspace{-0.3cm}
\caption{Comparison across different attention strategies as well as various decoder query initializations. }
\vspace{-0.3cm}
\label{tab:ablation_comparison_deformable_attention}
\end{table}

% Here we include the plot to comment about the offset distance. Mention that ours does not converge so close to the initialization, showing that it is actually capable of learning something "interesting". Also, ours show that it is capable of going beyond the boundaries of the action, which makes it more interesting to better refine the boundaries.

\subsection{Conditioning signal for the sparse stream $\mathcal{S}$}

To effectively integrate RDSA, it is imperative to obtain an expressive video-based conditioning signal --i.e., the red arrow in Fig.~\ref{fig:main_architecture}. 
As shown in Tab.~\ref{tab:performance_with_different_conditioning_signals}, using the raw video representation $\mathbf{V}^\ell$ significantly degrades performance, highlighting the need for a richer signal incorporating textual and temporal information. While conditioning on the outputs of the CA or SA modules of the dense stream $\mathcal{D}$ improves results, these lack non-linearities. Instead, using the non-linear PFFN output provides greater flexibility, leading to superior expressiveness. Notably, this choice also enhances HD performance, underscoring the benefits of task interaction between MR and HD.

% significantly decreases the tested metrics. This underpins the need for a richer signal that includes textual and temporal information. Defining the conditioning signal as the output of the CA module, or similarly, the output of the SA module of the dense stream $\mathcal{D}$, considerably improves the performance. Nevertheless, we conjecture that these still compare unfavorably to our chosen signal given their lack of non-linearities. We, in contrast, use the output of the non-linear PFFN that gives the model additional flexibility to learn more expressive conditioning signals. Interestingly, this choice significantly impacts HD as well, highlighting that the task interaction benefits both MR and HD.

\begin{table}
\centering
\resizebox{0.48\textwidth}{!}{
\begin{tabular}{l|ccccc|cc}
\toprule
\textbf{Type of features} & \multicolumn{5}{c}{\textbf{MR}} & \multicolumn{2}{c}{\textbf{HD}} \\

& \textbf{R1@0.5} & \textbf{R1@0.7} & \textbf{mAP@0.5} & \textbf{mAP@0.75}& \textbf{mAP} & \textbf{mAP} & \textbf{HIT@1} \\
\midrule
Raw. rep         & \underline{72.58} & 56.37 & 70.12 & 50.99 & 50.74 & 44.13 & \underline{71.37}          \\
Post CA          & 72.26 & \underline{58.39} & \underline{72.20} & 55.92 & 54.83 & 43.64 & 70.65           \\
Post CA+SA       & 72.13 & 58.32 & 72.00 & \underline{56.57} & \underline{54.94} & \underline{43.78} & 70.06           \\
\midrule
Post CA+SA+FFN   & \textbf{73.68} & \textbf{60.90} & \textbf{73.53} & \textbf{57.42} & \textbf{55.60} & \textbf{44.00} & \textbf{72.00} \\ 
\bottomrule
\end{tabular}}
\vspace{-0.2cm}
\caption{Evaluation of various conditioning signals.}
\vspace{-0.3cm}
\label{tab:performance_with_different_conditioning_signals}
\end{table}

\subsection{Effect of each module of the sparse stream $\mathcal{S}$}
In this ablation, we empirically evaluate the contribution of each of the components of $\mathcal{S}$. Concretely, we evaluate the influence of the CA, the SA, the RDSA, and the final PFFN. Observe in Tab.~\ref{tab:importance_of_each_of_the_submodules} that the initial textual injection CA seems to be particularly relevant for the R1 metric, while the reasoning SA module is important for the mAP. Moreover, as expected, the RDSA module is critical. This is to be expected, as without it, the queries become oblivious to the input video, leaving no other option but that of randomly guessing.  Again, the additional expressivity that the final PFFN brings via its non-linearities proves to be beneficial, despite affecting less than the other modules.

\begin{table}
\centering
\resizebox{0.48\textwidth}{!}{
\begin{tabular}{cccc|ccccc|cc}
\toprule
\textbf{CA} & \textbf{SA} & \textbf{RDSA} & \textbf{FFN} & \multicolumn{5}{c}{\textbf{MR}} & \multicolumn{2}{c}{\textbf{HD}} \\

& & & & \textbf{R1@0.5} & \textbf{R1@0.7} & \textbf{mAP@0.5} & \textbf{mAP@0.75} &\textbf{mAP} & \textbf{mAP} & \textbf{HIT@1} \\
\midrule
 & \checkmark & \checkmark & \checkmark & 71.42 & 57.74 & 71.73 & 56.18 & 54.47 & 43.53 & 71.35 \\
 \checkmark & & \checkmark & \checkmark & 71.87 & 58.71 & 72.13 & 55.99 & 54.32 & 43.28 & 69.55 \\
 \checkmark & \checkmark & & \checkmark & 59.68 & 30.71 & 59.02 & 24.6 & 29.09 & 43.58 & 70.39 \\
  \checkmark & \checkmark & \checkmark & & 72.84 & 59.29 & 73.13 & 56.91 & 55.72 & 43.85 & 72.58 \\
\midrule
  \checkmark & \checkmark & \checkmark & \checkmark         & \textbf{73.68} & \textbf{60.90} & \textbf{73.52} & \textbf{57.42} & \textbf{55.60} & \textbf{44.0} & \textbf{72.00} \\ 
\bottomrule
\end{tabular}}
\vspace{-0.2cm}
\caption{Importance of the modules of the sparse stream $\mathcal{S}$.}
\vspace{-0.5cm}
\label{tab:importance_of_each_of_the_submodules}
\end{table}

\vspace{-0.1cm}
\section{Conclusions}

In this work, we introduced the \textit{Sparse-Dense Side-Tuner}, the first proposal-free side-tuning method for VTG. Our dual-stream architecture jointly optimizes dense embeddings and \textit{recurrent decoder queries}, carefully adapting to the sparse and dense nature of MR and HD. We also identify an inherent context limitation of the Deformable CA --a key component of proposal-free architectures like ours-- which motivates our proposed alternative, the \textit{Reference-based Deformable Self-Attention}. Finally, we provide the first effective integration of InternVideo2 to a side-tuning framework, leading to a substantial performance boost. We evaluate our method on QVHighlights, TACoS, and Charades-STA, surpassing previous ST works and attaining either SOTA or near-SOTA, all while maintaining a minimal parameter count and a low memory overhead. We leave as future work studying the often overlooked problem of the cross-domain generalization capabilities of these methods.\\

\noindent \textbf{Acknowledgments:} This work has been partially supported by the Spanish project PID2022-136436NB-I00 and by ICREA under the ICREA Academia programme.

{\small
\bibliographystyle{ieee_fullname}
\bibliography{egbib}
}

% Appendix
\clearpage
\noindent {\Large \textbf{Supplementary Material}\par}
\vspace{0.2cm}
\appendix

\renewcommand*{\thesection}{\Alph{section}}
\renewcommand*{\thefigure}{\Alph{figure}}
\renewcommand*{\thetable}{\Alph{table}}
\setcounter{section}{0}
\setcounter{table}{0}
\setcounter{figure}{0}

In this supplementary material, we provide additional details and various ablation studies of the main contributions of our work. Concretely. Sec.~\ref{sec:implementation_details} describes the implementation details of our proposed SDST architecture and of our tested evaluation setups. Sec.~\ref{sec:extended_experimental_setup} presents a detailed description of the datasets used in our experiments: QVHighlights, TACoS, and Charades-STA. Sec.~\ref{sec:supp_objective_functions} expands on the objective functions used by SDST, covering the losses used for both Highlight Detection (HD), Moment Retrieval (MR), as well as alignment losses critical for effective side-tuning. Sec.~\ref{supp:details_deformable_attention} provides the necessary background related to the deformable attention mechanism. We then conduct a series of ablation studies such as Sec.~\ref{sec:supp_ablation_fair_comparison} that evaluates SDST over other relevant baselines under a fair comparison using only InternVideo2-1B features. Sec.~\ref{sec:ablation_efficiency_analysis} extends the efficiency analysis of the main text, analyzing multiple efficiency metrics like memory, parameters or running time. Sec.~\ref{sec:ablation_optimization_difficulty} ablates over the training stability of our proposal, and consequently the difficulty of its corresponding optimization.  Sec.~\ref{sec:comparison_other_petl_metl_methods} extends our comparison against other Parameter-Efficient and Memory-Efficient Fine-Tuning methods. Sec.~\ref{sec:parameter_sharing_supp} examines the impact of parameter sharing across intermediate refinement layers and Sec.~\ref{sec:ablation_interm_feats} studies the contribution of intermediate features when leveraging InternVideo2-1B. Additionally, Sec.~\ref{sec:supp_ablation_deformable_att} provides an in-depth analysis of our proposed RDSA mechanism, including the effect of different sampling strategies, the additional CNN-based context-enhancing module, and an extended analysis of the predicted offsets. Sec.~\ref{sec:ablation_ordering_sparse_stream} investigates the ordering of the modules in the sparse stream $\mathcal{S}$ and its impact on performance. Finally, Sec.~\ref{sec:supp_statistical_significance} presents various statistical significance tests, including Friedman and Nemenyi analyses, that confirm the robustness of our results.
\section{Implementation details}\label{sec:implementation_details}
In this section, we describe the most relevant implementation details of our proposed SDST architecture, a summary of which is also provided in Tab.~\ref{tab:model_config}. Note that the reported hyperparameters correspond only to the best-performing model. These hyperparameters remain mostly fixed w.r.t. to existing ST works like \cite{liu2024r} and only introduce a handful new hyperparameters which we set to $1.0$ for simplicity, while we do perform grid search to optimize the learning rate.

Overall, our method is implemented using PyTorch2.0 and CUDA12.8 and runs on a single NVIDIA RTX 6000 GPU with a precision of fp16. Our models are optimized, unless stated otherwise, using AdamW with a learning rate of $1e-4$ and a weight decay of $1e-4$. This follows a step-based schedule, decaying every 20 epochs. We apply linear warmup for the first 2000 iterations with a warmup ratio of 0.001 and clip gradients to a max norm of 35.

Our model operates with a hidden dimension of 256 and leverages a sinusoidal positional encoding. The entire model relies on ReLU non-linearities to enhance the model expressivity. To improve the regularization we use a dropout of 0.5 and incorporate a droppath with a drop probability of 0.25. Our model incorporates various Transformer blocks for instance for cross-modality injection and temporal relation learning, all of which have 8-heads, an attention dropout of 0.0, and an attention output dropout of 0.0. The attention modules are initialized with Xavier. The feedforward modules of the transformer block use a hidden dimension ratio of 4 times the chosen hidden dimension and also leverage a dropout of 0.0 and a Kaiming initialization. Importantly, following standard practices, we always incorporate residual connections to improve the stability of the training and follow a PostNorm strategy~\cite{vaswani2017attention} that normalizes the input based on a learnable LayerNorm module based on PostNorm.

Architecturally, our model consists of a dense and sparse stream, as well as their respective prediction heads. The dense stream incorporates various Transformer blocks for cross-modality injection and temporal relation learning, all of which have 8-heads, an attention dropout of 0.0, and an attention output dropout of 0.0. The attention modules are initialized with Xavier. The feedforward modules of the transformer block use a hidden dimension ratio of 4 times the chosen hidden dimension and also leverage a dropout of 0.0 and a Kaiming initialization. Importantly, throughout the Transformer blocks, we normalize the input through a learnable LayerNorm module based on PostNorm. Additionally, one of the key components of our Sparse stream is the use of our novel deformable attention mechanism named RDSA. This first applies a context-enhancing CNN is defined as a 2-layer CNN with a hidden dimension of 256, a learnable LayerNorm, and a non-linearity. Then, after concatenating the left-most, center, and right-most tokens, it uses an MLP to project them to a 64-dimension latent space. This is then used to apply two simple linear projections to compute the 4 different sampled Keys, with their respective attention scores. 

For the different prediction heads, we distinguish various different modules. On the one hand, the CLS and Regression heads are defined as a 1 and 3-layer MLPs, respectively. On the other hand, we define the actionness head following previous works like \cite{liu2022end}, which uses RoiPooling with a Roi size of 16. These roi features are then used for the actionness prediction, applying a 3-layer MLP. 

Finally, we make several important training considerations. All the experiments across the different datasets use a batch size of 32 and a minimum video length of 5. FPS is set to $0.5$ for QVHighlights and TACoS, and $1.0$ for Charades-STA.. We train for 60 epochs on QVHighlights, 50 on Charades-STA, and 150 on TACoS. The number of queries per sample varies on the nature of the dataset. In QVHighlights and Charades-STA, for instance, we define 30 different queries, while for TACoS we use only 5. For Charades-STA, we use a slightly higher learning rate of $2.5 \times 10^{-4}$ with a decay schedule of 30 epochs.

\begin{table*}
    \centering
    \small
    \begin{tabular}{c|c|c|c}
        \toprule
        \textbf{Pipeline component} & \textbf{Module} & \textbf{Field}  & \textbf{Value} \\
        \midrule
        
        \multirow{18}{*}{Architecture} & \multirow{15}{*}{General config} & Dropout & 0.5 \\
                                       & & K & 4 \\
                                       & & PE & Sinusoidal \\
                                       & & Hidden dimension & 256 \\
                                       & & Droppath & 0.25 \\
                                       & & Non-Linearities & ReLU \\
        
        & & FFN ratio & 4 \\
        & & Attention dropout & 0.0 \\
        & & FFN dropout & 0.0 \\
        & & Attention output dropout & 0.0 \\
        & & FFN output dropout  & 0.0 \\
        & & PreNorm & No \\
        & & Normalization type & LN \\
        & & Attention initialization & Xavier \\
        & & FFN initialization & Kaiming \\
        
        \cmidrule(lr){2-4}
        & \multirow{1}{*}{Sparse module} & Deformable sampling points & 4 \\

        \cmidrule(lr){2-4}
        & \multirow{2}{*}{CLS head} & Type & MLP \\
        & & Depth & 1 \\
        \cmidrule(lr){2-4}
        & \multirow{2}{*}{Regression head} & Type & MLP \\
        & & Depth & 3 \\
        \cmidrule(lr){2-4}
        & \multirow{4}{*}{Actionness head} & Type & MLP \\
        & & Depth & 3 \\
        & & Roi size & 16 \\
        & & Roi scale & 0 \\

        \midrule
        \multirow{9}{*}{Optimization} & Optimizer & & AdamW \\
        \cmidrule(lr){2-4}
        & Learning Rate & &  1e-4 \\
        \cmidrule(lr){2-4}
        & Weight Decay & & 1e-4 \\
        \cmidrule(lr){2-4}
        & \multirow{2}{*}{LR Schedule} & Type & Step-based \\
        & & Decay rate & Every 20 epochs \\
        \cmidrule(lr){2-4}
        & \multirow{3}{*}{Warmup strategy} & Type & Linear \\
        & & N. iterations & 2000 \\
        & & Ratio & 0.001 \\
        \cmidrule(lr){2-4}
        & Gradient clipping & Max norm & 35 \\

        \midrule
        \multirow{16}{*}{Datasets} & \multirow{4}{*}{QVHighlights} & Batch size & 32 \\
        & & FPS & 0.5 \\
        & & Min video len & 5 \\
        & & Epochs & 60 \\
        & & Num. queries & 30 \\
        \cmidrule(lr){2-4}
        & \multirow{7}{*}{Charades-STA} & Batch size & 32 \\
        & & FPS & 1.0 \\
        & & Min video len & 5 \\
        & & Epochs & 50 \\
        & & Num. queries & 30 \\
        & & Learning rate & 2.5e-4 \\
        & & Learning rate schedule & 30 \\
        \cmidrule(lr){2-4}
        & \multirow{5}{*}{TACoS} & Batch size & 32 \\
        & & FPS & 0.5 \\
        & & Min video len & 5 \\
        & & Epochs & 150 \\
        & & Num. queries & 5 \\

        \bottomrule
    \end{tabular}
    \caption{Summary of the most relevant hyperparameters and implementation details of our model.}
    \label{tab:model_config}
\end{table*}

\section{Description of the chosen datasets}\label{sec:extended_experimental_setup}

To test the effectiveness of our proposed \textit{SDST}, we conduct experiment on three different datasets --i.e., QVHighlights\cite{lei2021detecting}, TACoS~\cite{rohrbach2014coherent} and Charades-STA~\cite{gao2017tall}.
\paragraph{QVHighlights:} QVHighlights is the only dataset among the three that provides annotations of both MR and HD tasks. Concretely, this comprises 10k YouTube videos of humanly-annotated NLP queries of a vast variety of topics, from daily activities. For convenience, these videos are trimmed to a maximum duration of 150 seconds.

\paragraph{TACoS:} TACos is a widely used dataset for MR consisting of only 127 videos of cooking scenes with an average duration of 287 seconds. Overall, this includes 19k sentence-moment pairs. Notice that following previous works from the literature, we adapt this dataset to support our multi-task-based model by generating synthetic saliency annotations. For this, we consider a frame to have a saliency score of 1 if this belongs to the action, and 0 otherwise.

\paragraph{Charades-STA}:  Charades-STA extends the original Charades dataset, including 10k videos and 16k different sentence-moment annotations that capture a variety of indoor activities, making it a relevant benchmark to evaluate models in everyday human activity understanding. 

\section{Descriptions of the objective functions}\label{sec:supp_objective_functions}
In this section, we describe in greater detail the different objective functions that we used for our proposed SDST.

\subsection{Highlight detection loss}
% TALK ABOUT THE SALIENCY LOSS (SAMPLED NCE)
Given the dense visual embedding of the final refinement layer $\textbf{D}^K \in \mathbb{R}^{T \times F}$, we first apply a learnable AdaptivePooling mechanism to produce a single aggregated textual embedding $\textbf{T}^{pool} \in \mathbb{R}^{F}$ from the original textual representations $\textbf{T}^K \in \mathbb{R}^{L \times F}$. We then define the per-frame saliency scores $\hat{\textbf{Y}}^s \in \mathbb{R}^T$ as 
\begin{align}
    \small
    \hat{\textbf{Y}}^s = cos\_sim(\textbf{D}^K, \textbf{T}^{pool}) = \frac{\sum_{j=1}^{F} \textbf{D}^K_j \textbf{T}^{pool}_{j}}{||\textbf{D}^K|| \;|\textbf{T}^{pool}||}.
\end{align}
where this cosine similarity is computed for each of the visual frames. To ensure that a higher score corresponds to a higher relevance of a given frame w.r.t. the textual embedding $\textbf{T}^{pool}$ --corresponding to the last layer $K$-- we use a SampledNCE loss, which ranks the positive frames.

\subsection{Moment Retrieval losses}
Following the standard DETR pipeline, we apply the Hungarian algorithm to obtain a one-to-one matching between the predicted moment boundaries $\textbf{R}^\ell \in \mathbb{R}^{M \times 2}$ at each intermediate layer $\ell$, and the ground-truth (GT) annotations $\textbf{Y}^m \in \mathbb{R}^{M^* \times 2}$. Note that unless stated otherwise, we refer to the corresponding matches of the ground-truth $\textbf{Y}_m$ as $\hat{\textbf{Y}}^m \in \mathbb{R}^{M^* \times 2}$. Below we describe the different objective functions that we apply to these matching embeddings.

% TALK ABOUT THE FOCAL LOSS
\noindent \textbf{Classification loss:} The classification loss takes the predicted action probabilities of the M different recurrent decoder queries $\hat{\textbf{p}} \in \mathbb{R}^{M \times 1}$ and brings the probability of the unmatched proposals to $0$, while the rest have a probability of $1$. Given the imbalance between matched and unmatched queries, we leverage a FocalLoss
\begin{align}
\small
\mathcal{L}_{\text{cls}} = -\frac{1}{M} \sum_{m=1}^{M} \alpha (1 - \hat{\textbf{p}}_m)^\gamma \log (\hat{\textbf{p}}_m),
\end{align}
where  $\hat{\textbf{p}}_m$  is the predicted probability for proposal $m$ , and  $\alpha$  and  $\gamma$  are the standard Focal Loss hyperparameters that help address the class imbalance.

% TALK ABOUT THE L1 AND IOU LOSSES
\noindent \textbf{Regression losses:} We then focus our attention on the actual regression of the boundaries. For this, following previous DETR works~\cite{lei2021detecting} we first define an L1 loss given by
\begin{align}
\small
\mathcal{L}_{\text{L1}} = \frac{1}{M^*} \sum_{i=1}^{M^*} \left| \hat{\textbf{Y}}^m_i - \textbf{Y}^m_i \right|,
\end{align}
minimizing the absolute error between the predicted and ground-truth segment boundaries. Additionally, we employ an IoU-based loss \cite{liu2022end} to maximize the overlap between the predicted and GT action segments:
\begin{align}
\small
\mathcal{L}_{\text{IoU}} = 1 - \frac{\sum_{i=1}^{M^*} \text{IoU}(\hat{\textbf{Y}}^m_i, \textbf{Y}^m_i)}{M^*},
\end{align}
where  $\text{IoU}(\hat{\textbf{Y}}^m_i, \textbf{Y}^m_i)$  is the intersection-over-union between the predicted and ground-truth segments. 

% TALK ABOU THE GIOU LOSS FOR THE MAXIMUM
\noindent \textbf{Actionness losses:} As described in the Sec.~\ref{sec:prediction_head_and_training_objectives}, our NMS post-processing first considers the CLS score, which measures the probability of a predicted segment to be matched to a GT. As shown by \cite{liu2022end}, this is not enough as another pillar to an effective post-processing is having an estimate of the regression quality. For this, we also define the actionness scores $\hat{\textbf{Y}}^a \in \mathcal{R}^{M}$ as the maximum overlap of every query with any of the GT. In other words, for each of the learnable recurrent query embeddings we compute the maximum IOU with any of the GT actions. Then, we apply an L1 loss to regress this score which we can then leverage during inference. 

\begin{align}
\small
\mathcal{L}_{act} = \frac{1}{M} \sum_{i=1}^M \left| \hat{\textbf{Y}}^a_i - max_j^{M^*}(IOU(\textbf{R}^\ell_i, \textbf{Y}^m_j)) \right|
\end{align}
where $\hat{\textbf{\textbf{Y}}}^a_i$ is the predicted actionness score of the $i$-th recurrent decoder query, and $max_{j=1}^{M^*}(IOU(\textbf{R}^\ell_i, \textbf{Y}^m_j))$ is the maximum overlap of the $i$-th query w.r.t. any of the GT actions.

\subsection{Alignment losses}
% TALK ABOUT THE ROW AND COLUMNS SAMPLED NCE LOSSES
One critical aspect to guarantee an effective side-tuning is the inclusion of alignment losses which bring the visual and textual latent space closer in a semantically meaningful way. This is particularly important because, although the backbone has been pre-trained to ensure some degree of alignment, adapting it to a new domain such as VTG inevitably introduces domain shifts and noise. Without this alignment losses, this would hence significantly degree the quality of the features, and thus hinder the final performance. In our work we address this issue by introducing two contrastive losses \cite{liu2024r} that enforce video-query consistency at different levels of intermediate representation: 1) video-level alignment 2) layer-wise alignment. Notably, these losses are applied to all the intermediate layers independently.

\subsubsection{Video-level contrastive loss}
At a given level $\ell$, this loss enforces similarity between action-relevant frames and their corresponding textual query embedding. Specifically, it takes the embeddings $\textbf{V}^\ell$ and the pooled textual embedding $\textbf{T}^{pool}$ of that level, and pulls the positive frames --i.e., belonging to the action-- closer while pushing away the negatives. Interestingly, for a given $j$-th frame, we consider the negatives to be all the other $j$-th frames of the remaining batch elements, at the same refinement level $\ell$. Thereafter, we enforce our objective via an InfoNCE loss:
\begin{align}
\small
\mathcal{L}_{video\_cal} = InfoNCE(\textbf{V}^\ell, \textbf{T}^{pool})
\end{align}
where InfoNCE~\cite{oord2018representation} maximizes the similarity between the correct video-text pairs while promoting its separation from unrelated samples.

\subsubsection{Layer-wise contrastive loss}
This loss is similar to the previous one but operates across layers instead of the batch. This is, this ensures that the same frame-query pair learns different representations at two distinct levels $\ell$ and $\ell'$. This promotes that these representations are not redundant, and thus add complementary information to the model. To be more specific, following \cite{liu2024r} we define the negatives at level $\ell$ as the same frame-embedding but corresponding to a different intermediate layer $\ell'$.
\begin{align}
\small
\mathcal{L}_{video\_cal} = InfoNCE(\textbf{V}^\ell, \textbf{T}^{pool}).
\end{align}

\subsection{Inference}
During inference, we apply a soft NMS post-processing to filter out redundant action predictions. This algorithm sorts the proposals based on a confidence score, which in our case we define as the square root of the product of the class probabilities and actionness scores
\begin{align}
\small
\hat{\textbf{C}} = \sqrt{\hat{\textbf{p}} \cdot \hat{\textbf{Y}}^a}.
\end{align}
This prioritizes a high classification confidence together with a high localization confidence.

\section{Background on the deformable attention mechanism}\label{supp:details_deformable_attention}

The Vanilla Attention mechanism is the core component of Transformers, one of the most popular architectures in the community at the time of this writing. The attention mechanism can be defined as:
\begin{align}
    \small
    \mathbf{Q}=\mathbf{X}_\mathcal{Q} \mathbf{W}_\mathcal{Q} \;,\; \mathbf{K}=\mathbf{X}_\mathcal{K} \mathbf{W}_\mathcal{K}, \;\; \mathbf{V}=\mathbf{X}_\mathcal{V} \mathbf{W}_\mathcal{V}
\end{align}
\begin{align}
    \small
    \mathbf{S}= \frac{\sigma(\textbf{QK}^T)}{\sqrt{d_k}} \textbf{V}
\end{align}
Here $\sigma$ is a softmax activation, and $\mathbf{X}_\mathcal{Q}, \mathbf{X}_\mathcal{K}$ and $\mathbf{X}_\mathcal{V}$ define the inputs to the query, keys and values projection matrices $\mathbf{W}_\mathcal{Q}, \mathbf{W}_\mathcal{K}, \mathbf{W}_\mathcal{V}$, respectively. In the self-attention case, $\mathbf{X}_\mathcal{Q} = \mathbf{X}_\mathcal{K}$ while for cross-attention, $\mathbf{X}_\mathcal{Q} \neq \mathbf{X}_\mathcal{K}$.

This mechanism, despite very extended in the community these days, presents several important pitfalls like its quadratic complexity or its slow convergance. This motivated the proposal of various efficient attention mechanisms to attain similar performance while improving its efficiency. In this regard, we highlight the Deformable attention mechanism, proposed by \cite{zhu2020deformable}, inspired by the previous works on the deformable convolution. This mechanism attains a considerable efficiency boost with respect of the vanilla attention \cite{vaswani2017attention} by limiting the amount of attendable keys to a (small) predefined set of key tokens. More formally, the key of this module is the lack of explicit interaction between queries and keys, which are defined as
\begin{align}
    \small
    \mathbf{Q}=\mathbf{X}_\mathcal{Q}\mathbf{W}_\mathcal{Q}^{def} \;,\; \mathbf{K}=\mathbf{X}_\mathcal{K}\mathbf{W}_\mathcal{K}^{def},
\end{align}
where $\mathbf{W}_\mathcal{Q}^{def}$, $\mathbf{W}_\mathcal{K}^{def}$ are two linear projections. The deformable attention thus avoids computing the $\textbf{QK}^T$ similarity matrix, and instead employs an offset and attention-score predictors, $\mathcal{G}_\Delta$ and $\mathcal{G}_\textbf{A}$ to select and weight --only based on the queries-- a small subset $P$ of selectable keys. The output of this mechanism, the weighted aggregation of the selected keys, namely $\textbf{S}$, is computed as follows
\begin{align}
    \small
    \Delta= \mathcal{G}_\Delta(\mathbf{Q}) \in \mathbb{R}^{M \times P}, \; \mathbf{A} = \mathcal{G}_{\textbf{A}}(\mathbf{Q}) \in \mathbb{R}^{M \times P}, \\
    \mathbf{S} = \sum_{p=1}^P \left(\mathbf{A}_{:,p} \; \mathbf{K}[{\textbf{c} + \textbf{w} \odot \Delta_{:,p}}] \right) \in \mathbb{R}^{M\times F}
\end{align}
where $\mathcal{G}_\Delta$, $\mathcal{G}_\textbf{A}$ are often modeled as CNNs and $x[y]$ is the bilinear sampling of $x$ on index(es) $y$. This can be seen as a learnable method to identify, given a specific query, a small subset of key tokens to attend to, as well as their respective weight --necessary to compute their weighted average. The cornerstone of its efficiency boost is thus the fact that it does not require \textit{looking} at all the keys to make that decision --unlike Vanilla Attention-- but, instead, it is just a result of a simple projection layer of the query itself.

\section{Detailed ablation using only InternVideo2-1B features}\label{sec:supp_ablation_fair_comparison}

In order to guarantee a fair comparison between our proposed SDST and the rest of the tested baselines, in Tab.~\ref{tab:comparison_qvhighlights_with_internvideo2} we follow the work of \cite{gordeev2024saliency} and test the relevant baselines --on 3 independent seeds-- using only the InternVideo2-1B features. Note that we evaluate only the relevant baselines capable of doing MR and HD. Observe that our method substantially outperforms the other existing side-tuning method for VTG ~\cite{liu2024r}. Concretely, it improves its MR capabilities by a $3.82\%$ mAP, while it improves HD by a $2.21\%$ mAP and $2.1\%$ HIT@1. Similarly to our previous observations, our method outperforms the rest of the methods and remains competitive with the SG-DETR and even improves it in the two different HD metrics. We deem this to be particularly noteworthy given that our method poses only $27.3\%$ of the trainable parameters of SG-DETR.

% Table fair comparison QVHighlights with all InternVideo2
\begin{table}
\centering
\resizebox{0.48\textwidth}{!}{
\begin{tabular}{l|ccc|cc}
\toprule
\textbf{Method} & \multicolumn{3}{c|}{\textbf{MR-mAP}} & \multicolumn{2}{c}{\textbf{HD $\geq$ Very Good}} \\ 
 
\midrule
\multirow{1}{*}{M-DETR} & $60.20 \pm 0.55$ & $34.43 \pm 0.43$ & $35.40 \pm 0.41$ & $40.31 \pm 0.21$ & $63.89 \pm 0.62$ \\
\multirow{1}{*}{UniVTG}  & $63.51 \pm 0.25$ & $38.83 \pm 0.26$ & $37.78 \pm 0.16$ & $42.68 \pm 0.09$ & $69.34 \pm 0.23$ \\
\multirow{1}{*}{QD-DETR} & $67.78 \pm 0.29$ & $46.40 \pm 0.26$ & $45.52 \pm 0.15$ & $41.82 \pm 0.07$ & $68.06 \pm 0.24$ \\
\multirow{1}{*}{CG-DETR} & $69.86 \pm 0.21$ & $49.35 \pm 0.28$ & $48.69 \pm 0.17$ & $42.72 \pm 0.07$ & $69.87 \pm 0.15$ \\
\multirow{1}{*}{TR-DETR} & $70.08 \pm 0.15$ & $49.20 \pm 0.50$ & $47.99 \pm 0.42$ & $43.43 \pm 0.16$ & $71.13 \pm 0.25$ \\
\multirow{1}{*}{R2-Tuning\textsuperscript{*}} & $71.40 \pm 0.330$ & $53.786 \pm 0.684$ & $51.49 \pm 0.358$ & $41.72 \pm 0.085$ & $69.52 \pm 0.472$  \\
SG-DETR & \textbf{$73.52 \pm 0.05$} & \textbf{$57.91 \pm 0.13$} & \textbf{$55.64 \pm 0.20$} & \underline{$43.91 \pm 0.14$} & \underline{$71.47 \pm 0.73$} \\
\midrule
\textbf{Ours} & \underline{$73.20 \pm 0.226$} & \underline{$56.76 \pm 0.53$} & \underline{$55.31 \pm 0.23$} & \textbf{$43.93 \pm 0.063$} & \textbf{$71.62 \pm 0.348$}  \\
\bottomrule
\end{tabular}}
\caption{Evaluation of a set of representative baselines when leveraging InternVideo2-1b features, evaluated on QVHighlights \textit{val} split. \textbf{Bold} stands for the best and \underline{underline} for the second-best.}
\label{tab:comparison_qvhighlights_with_internvideo2}
\vspace{-0.2cm}
\end{table}

Additionally, in Tab.~\ref{table:comparison_charades_tacos_internvideo2_feats} we also show the analysis of the two remaining considered datasets, these being Charades-STA and TACoS. Similarly, in these two scenarios, our method improves the second-best performing works --i.e., FlashVTG for Charades-STA and SG-DETR for TACoS-- in all the metrics but R1@0.5 on TACoS where it incurs in a marginal degradation.

% Main results Charades-STA and TACoS
\begin{table}
\centering
\resizebox{0.38\textwidth}{!}{
\begin{tabular}{l|ccc|ccc}
\toprule
\textbf{Method} & \multicolumn{3}{c}{\textbf{Charades-ST}} & \multicolumn{3}{c}{\textbf{TACoS}} \\
& R1@0.5 & R1@0.7 & mIoU &  R1@0.5 & R1@0.7 & mIoU \\
\midrule
%M-DETR & & & & & &  \\
%TR-DETR & & & & & &  \\
R$^2$-Tuning & 68.2 & 46.26 & 58.14 & 38.02 & 25.27 & 35.36  \\
SG-DETR & 70.2 & 49.5 & 59.1 & \textbf{44.7} & \underline{29.9} & \underline{40.9} \\
FlashVTG & 70.3 & \underline{49.9} & -- & 41.8 & 24.7 & 37.6 \\

\midrule
\textbf{Ours} & \textbf{72.0} & \textbf{52.6} & \textbf{61.2} & \underline{44.5} & \textbf{32.3} & \textbf{42.2} \\
\bottomrule

\end{tabular}}

\vspace{-0.2cm}
\caption{Comparison of multiple representative baselines on Charades-STA and TACoS datasets when leveraging InternVideo2-1b features. \textbf{Bold} stands for the best and \underline{underline} for the second-best.}
\label{table:comparison_charades_tacos_internvideo2_feats}
\end{table}
\section{Additional efficiency analysis}\label{sec:ablation_efficiency_analysis}

In this paper, we normally study the efficiency of our model w.r.t. existing related works based only on the number of parameters. In Tab.~\ref{tab:extended_efficiency_stats} we extend this analysis to the training memory and the running time. For simplicity purposes, we limit this study to the QVHighlights dataset.

\begin{table}
\centering
\resizebox{0.42\textwidth}{!}{
\begin{tabular}{l|ccc}
\toprule
\textbf{} & \textbf{\# Params (M)} & \textbf{Memory (GB)} & \textbf{Runtime (it/s)} \\
\midrule
 Moment-DETR & 4.8  & 1.54 & 7.45  \\
 R2-Tuning & 2.7  & 2.4  & 5.55  \\
 TR-DETR & 7.9  & 1.76 & 4.75 \\
 HL-CLIP & 2.0  & 22.98 & 0.64  \\
 Llava-MR & 17.0  & $\approx 80 \times 8$  & -  \\
 MR.Blip & 19.0  & $\approx 80 \times 8$  & -  \\
 SG-DETR & 15.0 & - & - \\
 Flash-VTG & 10.9  & 2.3 & 5.2  \\
\midrule
 Ours & 4.1  & 3.4 & 4.16  \\
 \bottomrule
\end{tabular}}
\caption{Efficiency summary of a set of representative models evaluated on QVHighlight with InternVideo2-1b features, and a batch size of 32.}
\label{tab:extended_efficiency_stats}
\end{table}
\section{Study of the inherent optimization difficulty}\label{sec:ablation_optimization_difficulty}

Undeniably, existing SOTA models --e.g., \cite{liu2024r, gordeev2024saliency}-- accumulate a considerable number of losses and components. And unfortunately, ours is no exception. While this represents a considerable opportunity for future studies trying to create more compact models, in this section we aim to extend the ablation from Tab. \ref{tab:importance_of_each_of_the_submodules} of the main text where we justified the necessity of the various model components of the sparse stream $\mathcal{S}$. Concretely, first of all, we focus on the need for the different proposed losses --with the exception of those that are indispensable to solve the inherent task, like the saliency-related losses. For this, find in Tab. \ref{tab:importance_of_every_relevant_loss}.

\begin{table}
\centering
\resizebox{0.48\textwidth}{!}{
\begin{tabular}{ccccc|ccccc|cc}
\toprule
\textbf{$\mathcal{L}_1$} & \textbf{$\mathcal{L}_{IOU}$} & \textbf{$\mathcal{L}_{align}$} & \textbf{$\mathcal{L}_{act}$} & \textbf{$\mathcal{L}_{cls}$} & \multicolumn{5}{c}{\textbf{MR}} & \multicolumn{2}{c}{\textbf{HD}} \\

& & & & & \textbf{R1@0.5} & \textbf{R1@0.7} & \textbf{mAP@0.5} & \textbf{mAP@0.75} &\textbf{mAP} & \textbf{mAP} & \textbf{HIT@1} \\
\midrule

& \checkmark & \checkmark & \checkmark & \checkmark & 71.68 & 58.58 & 72.28 & 56.28 & 55.0 & 43.52 & 71.35 \\
\checkmark & & \checkmark & \checkmark & \checkmark & 73.94 & 59.55 & 72.07 & 55.21 & 54.25 & 43.35 & 71.26 \\
\checkmark & \checkmark & & \checkmark & \checkmark & 71.55 & 58.13 & 71.55 & 55.21 & 54.25 & 43.68 & 69.48 \\
\checkmark & \checkmark & \checkmark & & \checkmark & 73.1 & 59.61 & 73.87 & 57.01 & 55.6 & 43.29 & 71.00 \\
\checkmark & \checkmark & \checkmark & \checkmark & & 70.52 & 57.23 & 69.88 & 54.82 & 52.95 & 42.28 & 68.65 \\
\midrule
\checkmark & \checkmark & \checkmark & \checkmark & \checkmark & \textbf{73.68} & \textbf{60.90} & \textbf{73.52} & \textbf{57.42} & \textbf{55.60} & \textbf{44.00} & \textbf{72.00} \\

\bottomrule
\end{tabular}}
\caption{Importance of the main losses of our model when evaluated on QVHighlight val with InternVideo2-1b features.}
\vspace{-0.2cm}
\label{tab:importance_of_every_relevant_loss}
\end{table}

In terms of difficulty of optimization and parameter search, we highlight the considerable robustness in terms of hyperparameter choice. Concretely, as specified in Sec. \ref{sec:implementation_details}, the vast majority of the hyperparameters are kept consistent with previous relevant works like \cite{liu2024r}. The newly introduced hyperparameters were set to $1.0$ for simplicity. Nevertheless, this does not guarantee the robustness to this hyperparameter choice. Consequently we propose the following experiment: We randomly sample 4 additional different configurations --i.e., all the loss weights-- defining a range of $[0.25, 2]$ for $\lambda_{L1}, \lambda_{IOU}, \lambda_{act}$ and $\lambda_{cls}$ and from $[0.1, 0.5]$ for $\lambda_{sal}, \lambda_{align\_video}$ and $\lambda_{align\_layer}$. Observe in Tab. \ref{tab:permutation_results} the results of each of these configurations --defined in Tab. \ref{tab:configuration_of_hyperparameters}-- and observe that our model does not deviate significantly given these new random permutations. In fact, these more robust performance metrics --not requiring cherry picking the best configuration-- would still rank equally in the overall ranking from Tab. \ref{tab:main_results_qvhighlights}.

\begin{table}[t]
\centering
\resizebox{\linewidth}{!}{
\begin{tabular}{l|ccccc|cc}
\toprule
\textbf{Perm. \#} & \multicolumn{5}{c|}{\textbf{MR}} & \multicolumn{2}{c}{\textbf{HD}} \\

& \textbf{R1@0.5} & \textbf{R1@0.7} & \textbf{mAP@0.5} & \textbf{mAP@0.7} & \textbf{mAP} & \textbf{mAP} & \textbf{HIT@1} \\
\midrule
1 & 73.03 & 59.10 & 72.78 & 56.31 & 55.17 & 43.70 & 70.58 \\
2 & 73.94 & 59.35 & 73.92 & 57.26 & 55.87 & 44.30 & 72.58 \\
3 & 72.19 & 57.87 & 72.43 & 56.22 & 54.72 & 44.03 & 71.74 \\
4 & 72.97 & 58.77 & 73.13 & 56.84 & 55.58 & 44.52 & 71.68 \\
Chosen & 73.68 & 60.90 & 73.52 & 57.42 & 55.60 & 44.00 & 72.00 \\
\midrule
\textbf{Mean±Std} & 73.16 ± 0.61 & 59.19 ± 0.98 & 73.15 ± 0.52 & 56.81 ± 0.48 & 55.38 ± 0.40 & 44.11 ± 0.27 & 71.71 ± 0.65 \\
\bottomrule
\end{tabular}}
\caption{Performance across different permutations with main retrieval and detection metrics.}
\label{tab:permutation_results}
\end{table}

\begin{table}
\centering
\resizebox{0.42\textwidth}{!}{
\begin{tabular}{c|ccccccc}
\toprule
\textbf{Perm \#} & \textbf{$\lambda_{L1}$} & \textbf{$\lambda_{IOU}$} & \textbf{$\lambda_{sal}$} & \textbf{$\lambda_{align\_video}$} & \textbf{$\lambda_{align\_layer}$} &\textbf{$\lambda_{act}$} & \textbf{$\lambda_{cls}$} \\
\midrule
1 & 1.47 & 1.91  & 0.11  & 0.18  & 0.42  & 0.84 & 1.27 \\
2 & 1.43 & 0.41  & 0.17  & 0.33  & 0.37  & 0.3  & 0.29 \\
3 & 1.81 & 0.92  & 0.36  & 0.42  & 0.23  & 0.63  & 0.64 \\
4 & 0.4 & 1.24 & 0.31  & 0.1  & 0.16  & 1.13  & 0.64 \\
Chosen & 1 & 1  & 0.1  & 0.1  & 0.1  & 1  & 1 \\

\bottomrule
\end{tabular}}
\caption{The randomly chosen 4 different loss weight configurations and the final chosen configuration.}
\vspace{-0.2cm}
\label{tab:configuration_of_hyperparameters}
\end{table}
\section{Comparison with other PEFT and MEFT methods}\label{sec:comparison_other_petl_metl_methods}
In this section, we compare our proposed SDST against other relevant PEFT methods when leveraging InternVideo2-1B features for QVHighlights \textit{val} split. Importantly, we note that we were unable to evaluate relevant methods based on Adapters, LORA, or Prompt-based, due to severe computational limitations. Specifically, these methods require full backpropagation through the frozen backbone, exceeding the memory capacity of our NVIDIA RTX 6000. This underscores the importance of MEFT methods like ST. Furthermore, \cite{liu2024r} shows that these memory-expensive alternatives underperform over ST for VTG, allowing us to safely restrict the scope of this ablation to \textit{w/o Tuning}, and to relevant ST baselines --i.e., E$^3$VA~\cite{yin2024parameter}, LoSA~\cite{gupta2024losa}, LST~\cite{sung2022lst} and R$^2$-Tuning~\cite{liu2024r}. Among these, only R$^2$-Tuning is naturally suitable for a multi-modal setup like ours. Consequently, for a fair comparison, we made minimal modifications to adapt the other baselines to our setting.

Observe in Tab.~\ref{tab:performance_comparison_peft} that all these baselines poses a comparable number of trainable parameters, with the exception of LoSA~\cite{gupta2024losa} which has a slightly higher count. Similarly, all these methods show a very efficient memory usage, which as mentioned before, contrasts with other PEFT alternatives. We find that while all these tested baselines considerably improve the \textit{w/o Tuning} on MR, they perform quite similarly in terms of HD. Overall, our proposed SDST improves all these methods, with a especially significant boost on HD.

\begin{table}
    \centering
    \resizebox{0.48\textwidth}{!}{
    \begin{tabular}{l|ccccccc}
        \toprule
        \textbf{Method} & \textbf{\#Params} & \textbf{Memory} & \multicolumn{3}{c}{\textbf{MR}} & \multicolumn{2}{c}{\textbf{HD}} \\
        & \textbf{(M)} & \textbf{(GB)} & R1@0.5 & R1@0.7 & mAP & mAP & HIT@1 \\
        \midrule
        w/o Tuning & 2.70 & 2.35 & 66.97 & 51.10 & 46.19 & 41.45 & 67.23 \\
        \midrule
        E$^3$VA~\cite{yin2024parameter} & 2.57 & 2.96 & 68.97 & 53.16 & 47.68 & 41.04 & 68.13 \\
        LoSA~\cite{gupta2024losa} & 6.40 & \textbf{2.39} & 72.13 & 58.32 & 53.73 & 41.82 & 68.19 \\
        LST~\cite{sung2022lst}& \textbf{2.04} & 2.49 & 70.32 & 55.55 & 50.59 & 41.53 & 69.48 \\
        R$^2$-Tuning\cite{liu2024r} & 2.70 & 2.44  & 70.84 & 55.35 & 51.30 & 41.64 & 69.74\\
        \midrule
        \textbf{Ours} & 4.10 & 3.40 & \textbf{73.68} & \textbf{60.90} & \textbf{55.60} & \textbf{44.00} & \textbf{72.00} \\
        \bottomrule
    \end{tabular}}
    \caption{Performance comparison of different tuning methods on QVHighlights \textit{val} split. \textbf{Bold} stands for the best.}
    \label{tab:performance_comparison_peft}
\end{table}

In Tab. \ref{tab:performance_comparison_peft_charades_tacos} we include the homologous analysis for Charades-STA and TACoS datasets which shows similar results.

\begin{table}
    \centering
    \resizebox{0.48\textwidth}{!}{
    \begin{tabular}{l|ccc|ccc}
        \toprule

         & \textbf{R1@0.5} & \textbf{R1@0.7} & \textbf{mIOU} & \textbf{R1@0.5} & \textbf{R1@0.7} & \textbf{mIOU} \\
        \midrule
        w/o Tuning & 67.69 & 45.56 & 57.98 & 34.22 & 21.82 & 32.51 \\
        \midrule
        E$^3$VA~\cite{yin2024parameter} & 66.13 & 45.11 & 56.23 & 38.77 & 26.02 & 36.15 \\
        LoSA~\cite{gupta2024losa} & 67.69 & 45.16 & 57.42 & 38.54 & 24.49 & 35.71 \\
        LST~\cite{sung2022lst} & 68.2 & 46.26 & 58.14 & 38.02 & 25.57 & 35.26 \\
        R$^2$-Tuning\cite{liu2024r} & 69.25 & 46.67 & 58.69 & 39.54 & 27.37 & 36.27 \\
        \midrule
        \textbf{Ours} & \textbf{72.00} & \textbf{52.60} & \textbf{61.20}  & \textbf{44.50} & \textbf{32.30} & \textbf{42.20} \\
        \bottomrule
    \end{tabular}}
    \caption{Performance comparison of different tuning methods for MR on Charades-STA and TACoS. \textbf{Bold} stands for the best.}
    \label{tab:performance_comparison_peft_charades_tacos}
\end{table}

\section{Parameter sharing in SDST}\label{sec:parameter_sharing_supp}
Tab. \ref{tab:comparison_shared_vs_unshared} compares the performance of SDST with and without parameter sharing. This is, we evaluate if creating independent SG side-tuners for each of the $K=4$ different intermediate layers results in an improved performance on the MR and HD tasks.  Observe that this is not the case. Despite the additional $8.32$M parameters, unsharing the different side-tuning modules in fact results in a performance degradation in all the tested metrics. We hypothesize that sharing the same alignment module (see Eq.2) with the subsequent $\mathcal{L}_{align}$ loss, promotes that 1) embeddings share a unique latent space while 2) different layers focus on different semantics. This allows the sharing of the remaining modules, which we observed contributes to stabilizing the optimization, and thus, improve performance.
\begin{table}
\centering
\resizebox{0.48\textwidth}{!}{
\begin{tabular}{c|ccccc|cc|c}
\toprule
\textbf{Shared} & \multicolumn{5}{c}{\textbf{MR}} & \multicolumn{2}{c}{\textbf{HD}} & \textbf{Params} \\

& \textbf{R1@0.5} & \textbf{R1@0.7} & \textbf{mAP@0.5} & \textbf{mAP@0.75} &\textbf{mAP} & \textbf{mAP} & \textbf{HIT@1} \\
\midrule
\checkmark         & \textbf{73.68} & \textbf{60.90} & \textbf{73.52} & \textbf{57.42} & \textbf{55.60} & \textbf{44.0} & \textbf{72.00} & 4.10 M \\
                   & 71.23 & 57.1 & 71.62 & 55.44 & 54.1 & 43.82 & 70.45 & 12.43 M \\
\bottomrule

\end{tabular}}
\caption{Ablation of the effect of using shared vs unshared parameters on QVHighlights \textit{val} split. \textbf{Bold} stands for the best.}
\label{tab:comparison_shared_vs_unshared}
\end{table}

\section{Extended ablation on the use of intermediate InternVideo2-1B features}\label{sec:ablation_interm_feats}

In Sec.~\ref{sec:ablation_internvideo2_feats} we propose an ablation study to showcase the importance of the different refinement steps in performance, as well as quantify the effect of using intermediate layers instead of using the last-layer features only. For completeness, in Tab.~\ref{tab:ablation_num_intermediate_layers} we present the complete ablation with all the evaluated metrics that further support our findings and insights. 

\begin{table}
\centering
\resizebox{0.48\textwidth}{!}{
\begin{tabular}{l|c|ccccc|cc}
\toprule
\textbf{K} & \textbf{Interm?}  & \multicolumn{5}{c}{\textbf{MR}} & \multicolumn{2}{c}{\textbf{HD}} \\

& & \textbf{R1@0.5} & \textbf{R1@0.7} & \textbf{mAP@0.5} & \textbf{mAP@0.75}& \textbf{mAP} & \textbf{mAP} & \textbf{HIT@1} \\
\midrule
1 & \checkmark & 68.00 & 54.13 & 68.72 & 51.23 & 48.72 & 43.66 & 68.26        \\
2 & \checkmark & 71.94 & 57.48 & 71.98 & 55.15 & 53.64 & 43.49 & 69.61           \\
3 & \checkmark & 72.84 & 58.19 & 72.92 & 56.0 & 54.69 & 43.85 & 69.94           \\
4 & \checkmark & 73.68 & 60.90 & 73.52 & 57.42 & 55.60 & 44.0 & 72.00  \\
5 & \checkmark & 72.39 & 58.77 & 71.71 & 55.35 & 54.34 & 43.7 & 71.87           \\
\midrule
1 & & 68.00 & 54.13 & 68.72 & 51.23 & 48.72 & 43.66 & 68.26        \\
2 & & 73.23 & 59.29 & 72.75 & 56.61 & 55.06 & 44.18 & 71.68           \\
3 & & 73.61 & 59.42 & 73.27 & 56.41 & 55.32 & 43.69 & 70.84           \\
4 & & 70.84 & 57.03 & 72.02 & 54.27 & 54.6 & 44.29 & 71.55  \\
5 & & 70.00 & 57.03 & 70.62 & 54.72 & 53.54 & 43.27 & 68.52           \\
\bottomrule
\end{tabular}}
\caption{Ablation of the effect of refining over multiple refinement levels --i.e., last k-- and of the use of intermediate versus last-layer features. Results correspond to QVHighlights \textit{val} split}
\label{tab:ablation_num_intermediate_layers}
\end{table}
\section{Study of deformable attention}\label{sec:supp_ablation_deformable_att}

\noindent \textbf{Action-length-based analysis:} In Sec.~\ref{sec:ablation_deformable_attention_main_text} we expose the empirical benefits of our proposed RDSA method over the standard CA~\cite{vaswani2017attention}, Deformable CA~\cite{zhu2020deformable} and even decoder query initialization mechanisms like \cite{zhang2022dino}. To complement these results, in Tab.~\ref{tab:ablation_comparison_deformable_attention_maps} we disentangle the MR performance according to the action length --i.e., short, middle, and long actions. This comparison indicates that one of the core limitations of the standard CA module is its almost complete inability to correctly identify short actions. Observe that its performance over short actions degrades by $14.33\%$ and $15.07$ w.r.t..the Deformable CA and RDSA, respectively. We also observe that our method is especially effective at predicting long actions, improving by $2.84\%$ mAP w.r.t. the Deformable CA.

\begin{table}
\centering
\resizebox{0.48\textwidth}{!}{
\begin{tabular}{l|cccc}
\toprule
\textbf{Att. strat.} & \textbf{mAP short} & \textbf{mAP middle} & \textbf{mAP long} & \textbf{mAP} \\
\midrule
Stand. CA & 3.31 & 45.92 & 51.17 & 42.72 \\

Def.CA & 17.64 & 57.68 & 56.92 & 54.27 \\
\midrule
Ours & \textbf{18.38} (+0.74) & \textbf{58.15} (+0.47) & \textbf{59.76} (+2.84) & \textbf{55.60} (+1.33) \\
\bottomrule
\end{tabular}}
\caption{Performance comparison of different attention strategies across different video lengths from the QVHighlights \textit{val} split. We include the absolute difference between our method and the second-best performing baseline --i.e., Def. CA. \textbf{Bold} stands for the best.}
\label{tab:ablation_comparison_deformable_attention_maps}
\end{table}

\noindent\textbf{Effect of the CNN and the sampling points:} In this section, we are also interested in providing further insights necessary for a deep understanding of RDSA. To be more specific, we focus our attention on two important aspects of this module, namely the points that are sampled to form the alternative query embeddings, and the additional CNN module to gain local context (see Eq.~\ref{eq:rdsa_new_queries}).

\begin{table*}
\centering
\resizebox{0.6\textwidth}{!}{
\begin{tabular}{c|c|ccccc|cc}
\toprule
\textbf{Sampling points} & \textbf{CNN} & \multicolumn{5}{c}{\textbf{MR}} & \multicolumn{2}{c}{\textbf{HD}} \\

& & \textbf{R1@0.5} & \textbf{R1@0.7} & \textbf{mAP@0.5} & \textbf{mAP@0.75} &\textbf{mAP} & \textbf{mAP} & \textbf{HIT@1} \\

\midrule

\multirow{2}{*}{c} &  & 71.94 & 58.39 & 71.37 & 55.35 & 53.57 & 43.59 & 71.23   \\
                   & \checkmark & 70.97 & 57.03 & 71.45 & 55.18 & 54.01 & 43.44 & 70.58          \\
\midrule
\multirow{2}{*}{l-r} &  & 72.26 & 57.16 & 71.77 & 55.03 & 53.55 & 43.26 & 70.71 \\ 
                     & \checkmark & 72.32 & 57.74 & 72.31 & 55.91 & 54.36 & 43.28 & 71.81 \\  
\midrule
\multirow{2}{*}{l-c-r} &  & 72.13 & 58.77 & 71.54 & 54.81 & 53.94 & 43.25 & 71.68          \\
                       & \checkmark & \textbf{73.68}& \textbf{60.90} & \textbf{73.53} & \textbf{57.42} & \textbf{55.60} & \textbf{44.00} & \textbf{72.00}        \\

\bottomrule

\end{tabular}}
\caption{Ablation of the effect of different sampling strategies like center-sampling (c), left-most and right-most action-boundary sampling (l) and (r), respectively. We also quantify the importance of our additional CNN module for enhanced context learning. Results correspond to QVHighlights \textit{val} split. \textbf{Bold} stands for the best.}
\label{tab:ablation_comparison_deformable_attention_sampling_and_cnn}
\end{table*}

For this, in Tab.~\ref{tab:ablation_comparison_deformable_attention_sampling_and_cnn} we ablate over three possible sampling strategies. The first samples only the center frame of the action, the second samples both the left and right-extremum of the action boundaries, and the later samples all these 3 embeddings. Note that as described in Sec.~\ref{sec:rdsa_main_text} , these embeddings are sampled based on the predicted action reference, and after concatenation, are used as alternative query embeddings for a Deformable Self-Attention mechanism. 

In this regard, our ablation indicates that the RDSA benefits the most from the extremum embeddings when also incorporating a CNN module. This indicates that the CNN is effectively gathering context of the neighborhood of the current action boundaries, providing critical information for the offset prediction, and thus, of \textit{where the model should look} to further refine the predicted segments. 

We also observe that the center embeddings are necessary even though they seem to play a lesser role in the overall performance. Interestingly, the use of a CNN in fact harms the effectiveness of these embeddings. We conjecture that by definition, the center embeddings tend to be \textit{surrounded} by very action-like embeddings. Thus, the local neighborhood does not necessarily provide useful information, and might even cause learning instabilities or aggravate the overfitting.

In short,  these experiments show the importance of using the 3 proposed sampled embeddings, and the overall positive impact that the additional CNN module has on gathering information on the local neighborhoods of the action boundaries.

\begin{figure}
    \centering
    \begin{subfigure}{0.3\textwidth} % 1/3 of the width
        \centering
        \includegraphics[width=\textwidth]{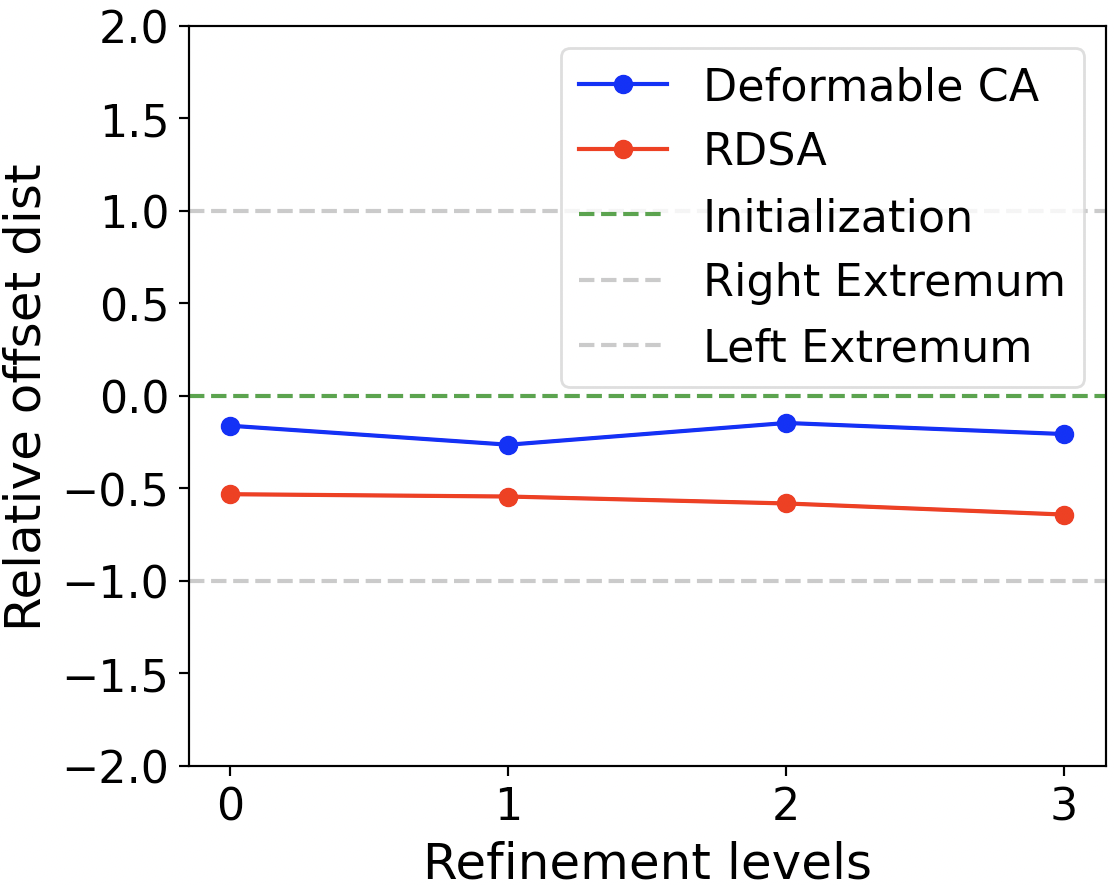}
    \end{subfigure}
    \caption{Attention-weighted offset distances across refinement levels over head 2 when evaluating QVHighlights \textit{val} split.}
    \label{fig:deformable_comparison_supp}

\end{figure}

\noindent \textbf{Extended analysis on the predicted offsets:} Finally, we extend the analysis provided in Sec.~\ref{sec:ablation_deformable_attention_main_text} that sheds light on where the predicted offsets point to. Concretely, in Fig.~\ref{fig:deformable_comparison_supp} we additionally depict a similar analysis for a head that initializes the heads near the center of the action --i.e., 0. Observe that in this case, similar to our previous observations, we find that the original Deformable CA~\cite{zhu2020deformable} keeps the offsets closer to the original initialization, suggesting a lack of proper understanding of the input video. Our method, in contrast, learns to point the offsets over the frames closer to the left-most boundary. Also, we find that in our model learns to \textit{look} more left as the model processes deeper levels.

% Here talk about the full ablation of the offset distances

\section{Study of the ordering of the different modules of the sparse stream}\label{sec:ablation_ordering_sparse_stream}

One important aspect to consider is the ordering of the 4 different modules of the sparse stream. For this, in Tab.~\ref{tab:comparison_ordering_submodules_sparse} we evaluate multiple relevant combinations. Note that we avoid ablating over the final FFN module due to computational limitations. In this regard, Tab.~\ref{tab:comparison_ordering_submodules_sparse} indicates that it is beneficial to include the CA module to gain textual context as early as possible.

\begin{table}
\centering
\resizebox{0.48\textwidth}{!}{
\begin{tabular}{l|ccccc|cc}
\toprule
\textbf{Mod. permutation} & \multicolumn{5}{c}{\textbf{MR}} & \multicolumn{2}{c}{\textbf{HD}} \\

& \textbf{R1@0.5} & \textbf{R1@0.7} & \textbf{mAP@0.5} & \textbf{mAP@0.75} &\textbf{mAP} & \textbf{mAP} & \textbf{HIT@1} \\
\midrule
SA-CA-Def-FFN         & 70.77 & 56.13 & 71.54 & 54.66 & 53.81 & 43.36 & 70.39 \\
SA-Def-CA-FFN         & 70.97 & 57.29 & 71.30 & 54.72 & 53.74 & 43.06 & 69.87 \\
Def-SA-CA-FFN         & 71.81 & 57.61 & 71.61 & 55.05 & 53.91 & 43.16 & 70.19 \\
Def-CA-SA-FFN         & 72.77 & 58.71 & 72.78 & 56.46 & 54.98 & 44.04 & 71.35 \\
CA-Def-SA-FFN         & 72.58 & 59.42 & 72.78 & 57.26 & 54.94 & 43.46 & 70.77 \\
\midrule
CA-SA-Def-FFN         & \textbf{73.68} & \textbf{60.90} & \textbf{73.52} & \textbf{57.42} & \textbf{55.60} & \textbf{44.00} & \textbf{72.00} \\
\bottomrule

\end{tabular}}
\caption{Ablation on the importance of the ordering of the components in the sparse stream when evaluated on QVHighlights \textit{val} split.}
\label{tab:comparison_ordering_submodules_sparse}
\end{table}

\section{Ablation statistical significance tests}\label{sec:supp_statistical_significance}

\begin{table*}
\footnotesize
\centering
\label{table:comparison-results}
\resizebox{0.9\textwidth}{!}{
\begin{tabular}{c|ccccc|cc|ccc|ccc}
\toprule

\multirow{4}{*}{\textbf{Method}} & \multicolumn{7}{c|}{\textbf{QVHighlights(test)}} & \multicolumn{3}{c|}{\textbf{Charades-ST}} & \multicolumn{3}{c}{\textbf{TACoS}} \\

\midrule

& \multicolumn{5}{c|}{\textbf{MR}} & \multicolumn{2}{c|}{\textbf{HD}} & & & & & &  \\

\midrule

 & \multicolumn{2}{c}{\textbf{R1}} & \multicolumn{3}{c|}{\textbf{mAP}} & \multicolumn{2}{c|}{\textbf{$\ge$ Very good}} & \multicolumn{2}{c}{\textbf{R1}} & & \multicolumn{2}{c}{\textbf{R1}} & \\
 
 & \textbf{@0.5}  & \textbf{@0.7} & \textbf{@0.5} & \textbf{@0.75} & \textbf{Avg.} & \textbf{mAP} & \textbf{HIT@1} & \textbf{@0.5}  & \textbf{@0.7} & \textbf{mIOU} & \textbf{@0.5}  & \textbf{@0.7} & \textbf{mIOU} \\
\midrule
Moment-DETR & 9 & 9 & 9 & 9 & 9 & 7 & 6 & 9 & 9 & -- & 9 & 9 & 9 \\
QD-DETR & 7 & 7 & 7 & 7 & 7 & 5 & 4 & 8 & 8 & -- & 7 & 7 & 7 \\
UniVTG & 8 & 8 & 8 & 8 & 8 & 6 & 5 & 7 & 7 & 6 & 8 & 8 & 8 \\
CG-DETR & 5 & 6 & 5 & 6 & 6 & 4 & 2 & 6 & 6 & 5 & 5 & 6 & 5 \\
BAM-DETR & 6 & 5 & 5 & 5 & 5 & -- & -- & 5 & 4 & 4 & 4 & 3 & 3 \\
R2-Tuning & 4 & 4 & 4 & 4 & 4 & 3 & 3 & 4 & 5 & 3 & 6 & 4 & 6 \\
SG-DETR\textsuperscript{\textdagger} & 1 & 1 & 1 & 1 & 1 & 1 & 1 & 3 & 4 & 2 & 1 & 2 &2 \\
Flash-VTG\textsuperscript{\textdagger} & 3 & 3 & 2 & 3 & 3 & -- & -- & 2 & 2 & -- & 3 & 5 & 4 \\

\midrule
\textbf{Ours}\textsuperscript{\textdagger} & 2 & 2 & 3 & 2 & 2 & 2 & 1 & 1 & 1 & 1 & 2 & 1 & 1 \\
\bottomrule
\end{tabular}}
\caption{Comparison with the SOTA on QVHighlights \textit{test} and \textit{val}. Also note that for comparability purposes, none of these results rely on pre-training. {}\textsuperscript{\textdagger} indicates that the method uses InternVideo2 backbone (comparable to ours).} \label{tab:main_results_ranking}
\vspace{-0.3cm}
\end{table*}

In this section, we aim to assess if the performance of our proposed SDST (ours) significantly differs from the other relevant baselines. For our main results, we were unable to establish a fair comparison with the other baselines across various seeds, given that for instance QVHighlights \textit{test} has a limited number of submissions. Consequently, we focus on the statistical study of the different model rankings across all the 3 studied datasets and their respective metrics. Concretely, we carry out two primary statistical tests: the Friedman test~\cite{friedman1937use} and Nemenyi’s test~\cite{huber1967fifth}.

\subsection{Friedman Test}

The Friedman test is a non-parametric statistical test to test if $k$ different variables are part of the same population. Concretely, we apply this test to study if given a set of various models, their rankings differ significantly across different datasets and metrics.  We define the null hypothesis of the Friedman test as \textit{all models perform similarly}, hence implying that there are no significant differences in the rankings across datasets/metrics. 
Mathematically, the Friedman statistic  $\chi^2_F$  is given by

\begin{equation}
\chi^2_F = \frac{12}{N \cdot k \cdot (k+1)} \sum_{i=1}^{k} \left( R_i - \frac{N(k+1)}{2} \right)^2,
\end{equation}

where $N$  is the number of datasets and their respective metrics. $K$ is the number of tested models, and $R_i$ is the sum of ranks for model $i$ across the different datasets and metrics. 

In our case, the Friedman test yielded a statistic of  $\chi^2_F = 5.640$  with a p-value of $0.933$. This is greater than 0.05, the threshold that is typically employed to determine the statistical significance. Hence, we can reject the null hypothesis, and conclude that there is no significant difference between the rankings of the models evaluated on all datasets. In other words, we observe that the performance of the various evaluated baselines, including our SDST, perform consistently across different datasets and metrics.

\subsection{Pairwise Nemenyi’s Test}

In this second statistical significance test we are interested in a more fine-grained analysis that might allow us to determine if our proposed method performs significantly better than the remaining considered baselines --especially of the R2-Tuning and the SG-DETR. For this, we proceed with a pairwise comparison using the Nemenyi's test. More in detail, for each pair of models, the Nemenyi test statistic is calculated based on their respective rank differences across the various datasets and metrics. We present the obtained p-values in Tab.~\ref{tab:statistical_significance_test_pairwise}. These results indicate that our method (SDST) performs statistically better than all the other baselines with the exception of SG-DETR which performs statistically on par. This matches our previous observations and certifies that our method attains statistically equivalent performance to SOTA while using only $27\%$ of its respective parameter count.

\begin{table}
\footnotesize
\centering
\resizebox{0.48\textwidth}{!}{
\begin{tabular}{c|c|c}
\toprule
\textbf{Comparison w.r.t. SDST} & \textbf{p-value} & \textbf{Statistically different} \\
\midrule
\text{Moment-DETR} & 0.0012 & \checkmark \\
\text{QD-DETR} & 0.0013 & \checkmark \\
\text{UniVTG} & 0.0011 & \checkmark \\
\text{CG-DETR} & 0.0013 & \checkmark \\
\text{BAM-DETR} & 0.0013 & \checkmark \\
\text{R2-Tuning} & 0.0013 & \checkmark \\
\text{SG-DETR} & 0.7926 & \\
\text{Flash-VTG} & 0.0011 & \checkmark  \\
\bottomrule
\end{tabular}}
\caption{Nemenyi's significance test across the various pair-wise comparisons w.r.t. to our proposed SDST.}
\label{tab:statistical_significance_test_pairwise}
\end{table}

These results indicate that for Ours vs SG-DETR, the p-value is 0.7926, meaning there is no statistically significant difference between the two methods. In contrast, for comparisons between Ours and the other models, the p-values are all below 0.05, suggesting that Ours is significantly better than the other models.

\end{document}